\RequirePackage{fix-cm}
\documentclass[10pt,twocolumn,letterpaper]{article}
\usepackage[pagenumbers]{cvpr}
\usepackage{placeins,array,booktabs,amsmath,amssymb,graphicx,alphalph,tabularx}
\usepackage{chapterbib}

\definecolor{cvprblue}{rgb}{0.21,0.49,0.74}
\usepackage[pagebackref,breaklinks,colorlinks,allcolors=cvprblue]{hyperref}
\title{OSEF: One-Step Evidence Fusion for Cross-Video Scene Procedure Planning}
\author{Zhentong Ye \quad Lei Zhang \quad Sijia Zhou \quad Yingda Yu \quad Yuehan Shi\\
Jiaqi Xuan \quad Shuaiwu Dong \quad Guanchao Tong\textsuperscript{*} \quad Meimei Zhang \quad Bin Li\textsuperscript{*}\\
{\small \textsuperscript{*}Corresponding authors}}

\hypersetup{pdfauthor={Zhentong Ye, Lei Zhang, Sijia Zhou, Yingda Yu, Yuehan Shi, Jiaqi Xuan, Shuaiwu Dong, Guanchao Tong, Meimei Zhang, Bin Li}}
\makeatletter
\patchcmd{\@maketitle}{\vspace*{24pt}}{\vspace*{12pt}}{}{\PackageError{r18layout}{Title spacing patch failed}{}}
\makeatother
\input{platform_setup280}

\begin{document}
\input{platform_refs280}
\PlatformMainPhase
\widowpenalty=10000

\maketitle
\begin{abstract}
Video Scene Procedure Planning (VSPP) predicts actions between supplied start and goal observations. We introduce Cross-Video Scene Procedure Planning (CVSPP), which instead takes a start--goal query and candidate videos, and predicts an action sequence together with its supporting video and interval. Two challenges arise: demonstrations of the same task can support different transitions, and an early retrieval error can deprive the planner of relevant evidence. We construct a dedicated benchmark from eleven video sources, with task-specific queries, candidate sets, and linked video--interval--plan references. Its fourteen settings support a broad study of nine adapted planner families. We propose One-Step Evidence Fusion (OSEF), which retains candidate features while producing interval-conditioned planning representations. Separate boundary scores model the supporting interval, and compatible-mass decoding combines the probability of the same complete plan across overlapping intervals before selecting the output. Joint evaluation requires video, interval, and plan to match one reference example. Across four matched development-data comparisons on NIV, CrossTask, and COIN, compatible-mass decoding increases mean joint success by $2.40$--$6.79$ percentage points without retraining. On NIV, query-only controls retain plan accuracy yet sharply reduce joint success; donor-interface studies examine how VSPP models use retrieved visual states. We further identify temporal-reference sensitivity: changing only the reference projection can reverse the decoding gain, exposing an evaluation challenge for the new task.

\end{abstract}
\section{Introduction}
Procedure planning in instructional videos predicts the actions that connect a start state to a goal state~\cite{procplan2020}.
For example, a planner may infer \texttt{chop}, \texttt{saut\'e}, and \texttt{plate} between raw ingredients and a finished dish.
Video Scene Procedure Planning (VSPP) studies this problem with supplied visual observations of the two endpoints~\cite{p3iv2022,pdpp2023,schema2024,planllm2025}.
A video collection offers another useful setting: a user describes the transition and wants both the intervening steps and a recording that demonstrates them.
The supporting recording and its relevant interval must then be identified together with the plan.

\begin{figure}[t]
\centering
\includegraphics[width=\columnwidth]{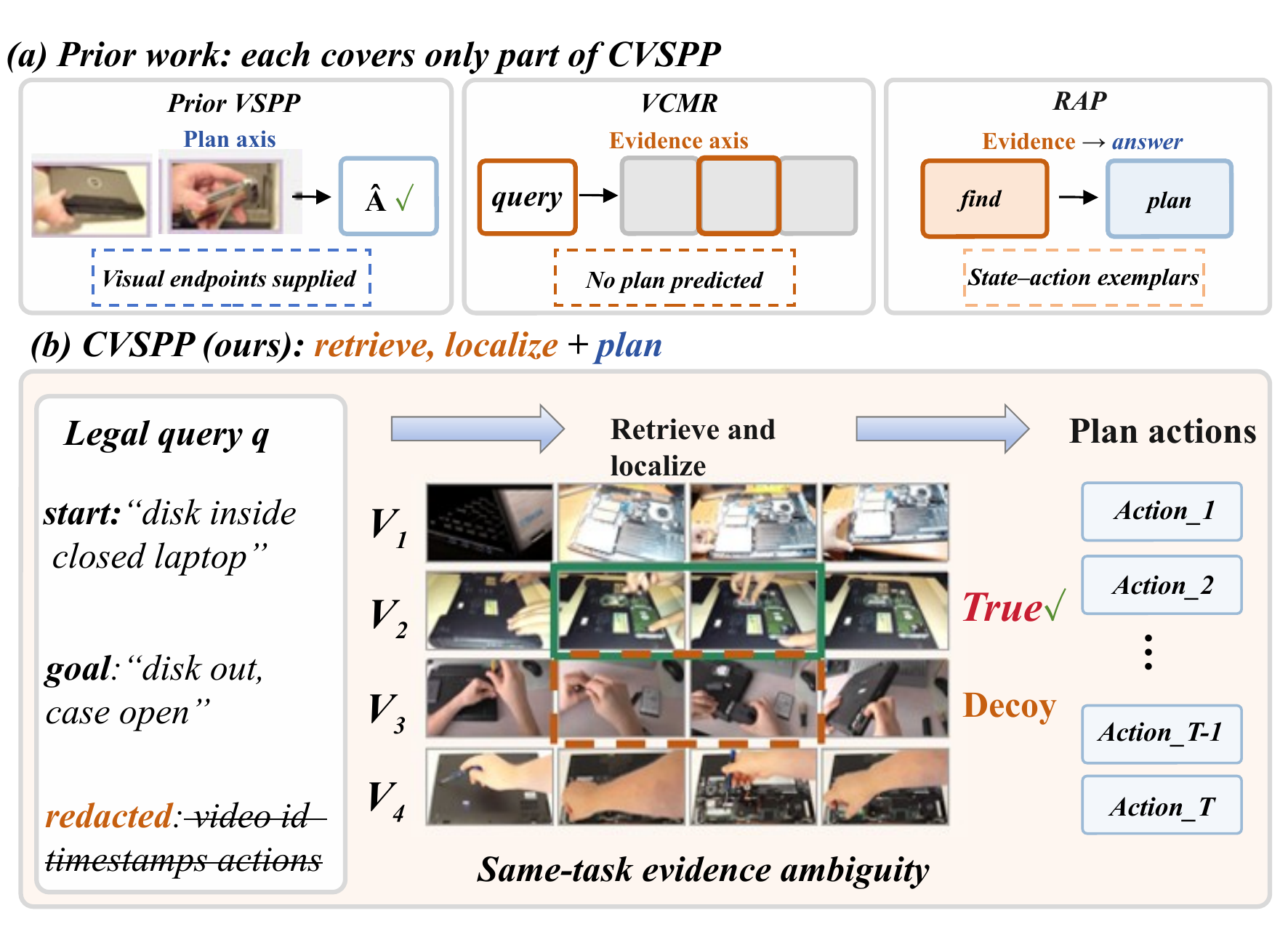}
\caption{\textbf{CVSPP links actions to video evidence.} VSPP uses supplied endpoints; video-corpus moment retrieval (VCMR) locates a given description; RAP retrieves exemplars. CVSPP predicts unknown actions and their supporting interval. Same-task candidates illustrate the ambiguity.}
\label{fig:task}
\end{figure}

We introduce \textbf{Cross-Video Scene Procedure Planning (CVSPP)} for this setting.
Given a start--goal query, a finite set of candidate videos, and a requested plan length, the model returns a video, an interval, and an action sequence (Figure~\ref{fig:task}).
Two challenges connect these outputs.
First, \textbf{same-task evidence ambiguity} makes a task name insufficient: demonstrations of the same procedure can contain different stages and neighboring windows.
The query must specify the requested transition, and the model must associate its plan with the corresponding support.
Second, \textbf{retrieval-to-plan error propagation} arises when a planner receives only an incorrectly selected recording or crop.
Keeping useful evidence available until a complete output is considered provides an alternative to this early commitment.

CVSPP brings together capabilities studied separately in procedure planning, video-corpus moment retrieval, and retrieval-augmented planning~\cite{procplan2020,conquer2021,rap2024}.
Its distinction is the linked output: the intermediate action sequence is unknown, yet it must agree with the selected video interval.
Candidate recordings can expose intermediate observations absent from endpoint-only VSPP.
CVSPP therefore evaluates whether a predicted procedure is supported by a retrieved demonstration, with a different input and output contract.

To study this problem, we construct a benchmark from eleven video sources.
The construction supplies endpoint queries, candidate sets, and complete video--interval--plan references.
It supports fourteen source--plan-length settings and adaptations of nine established planner families.
We evaluate the components separately and require them to match \emph{the same} annotated example for joint success.
This distinction matters: a correct plan and a plausible interval need not belong to one valid complete output.

We propose \textbf{One-Step Evidence Fusion (OSEF)} to connect evidence selection and planning.
OSEF scores possible start and goal boundaries, retains candidate clip features, and uses an evidence-to-plan (E2P) adapter to form planning representations conditioned on each interval.
We further introduce compatible-mass (CM) decoding: overlapping intervals contribute their probabilities of the same complete plan before the final video--interval--plan decision.
The interface also supports adaptations of VSPP models by supplying predicted visual states in place of their original endpoint inputs.

Our contributions are fourfold.
\textbf{(1)} We formulate CVSPP and joint evaluation of its three linked outputs.
\textbf{(2)} We construct an eleven-source benchmark and study transfer through nine planner families across fourteen settings.
\textbf{(3)} We develop OSEF's evidence-to-plan interface, interval-conditioned formulation, and compatible-mass decision rule.
\textbf{(4)} We analyze what correct support requires through input controls, interface studies, and matched decoding comparisons, and identify temporal-reference sensitivity as a challenge for evaluating the new task.
The benchmark annotations, construction configurations, and evaluation code will be released upon acceptance.

\section{Related Work}
\paragraph{Video Scene Procedure Planning.}
VSPP predicts an action sequence between visual endpoints~\cite{procplan2020}.
Its methods provide complementary planning mechanisms: P3IV models probabilistic plans~\cite{p3iv2022}; SCHEMA represents state changes~\cite{schema2024}; PDPP, ActionDiffusion, and MTID use diffusion-based prediction~\cite{pdpp2023,actiondiffusion2025,mtid2025}; KEPP incorporates procedural knowledge~\cite{kepp2024}; and PlanLLM refines plans with a language model~\cite{planllm2025}.
Skip-Plan and ViterbiPlanNet exploit action structure~\cite{skipplan2023,viterbiplannet2026}.
Recent language-aware planning transforms visual endpoints into language representations~\cite{shi2026lap}, while RECIPE uses training-time grounding rewards for procedural continuation~\cite{seminara2026recipe}.
Table~\ref{tab:benchmark-results-r13}(a) adapts nine VSPP planners with predicted visual states.

\paragraph{Video Corpus Moment Retrieval.}
CONQUER studies locating a described moment in a video corpus~\cite{conquer2021}.
QD-DETR learns query-conditioned video representations~\cite{qddetr2023}; QVHighlights benchmarks moment retrieval and highlight detection~\cite{qvhighlights2021}.
HiREST combines moment retrieval with step segmentation and captions~\cite{hirest2023}; VERIFIED tests fine-grained corpus moment retrieval~\cite{verified2024}. Temporal-grounding work also studies annotation uncertainty~\cite{zhou2021uncertainty}.
These methods motivate the evidence side of CVSPP.
The additional requirement here is to infer an unknown action sequence and associate it with the returned support, rather than localizing a given text description.

\paragraph{Retrieval-Augmented Planning.}
RAP retrieves state--action exemplars from an external memory for procedure planning~\cite{rap2024}.
The retrieved objects differ from the query-time candidate videos and intervals returned by CVSPP.
Table~\ref{tab:task_positioning} summarizes these task interfaces.
For the final complete-output decision, our compatible-mass formulation builds on loss-aware prediction and probability-mass aggregation~\cite{nowozin2014optimal,ahmed2015optimizing}; its compatibility condition links video identity, temporal overlap, and the entire plan.

\begin{table}[t]
\centering\small
\begin{tabular}{@{}>{\raggedright\arraybackslash}p{.26\columnwidth}>{\raggedright\arraybackslash}p{.30\columnwidth}>{\raggedright\arraybackslash}p{.34\columnwidth}@{}}
\toprule
Setting & Supplied input & Requested output\\\midrule
VSPP~\cite{procplan2020} & Visual start and goal & Action sequence\\
VCMR~\cite{conquer2021} & Text and videos & Video and interval\\
RAP~\cite{rap2024} & Visual endpoints and exemplar memory & Variable-length action sequence\\
CVSPP & Start--goal text, videos, plan length & Video, interval, action sequence\\
\bottomrule
\end{tabular}
\caption{\textbf{Related tasks request different outputs.} Inputs and outputs are compared at inference. CVSPP requires its three outputs to match one complete reference example.}
\label{tab:task_positioning}
\end{table}

\section{CVSPP Benchmark Construction}
\label{sec:benchmark}
\begin{figure*}[t]
\centering
\begin{minipage}[t]{.60\textwidth}
\vspace{0pt}\centering
\includegraphics[width=\linewidth]{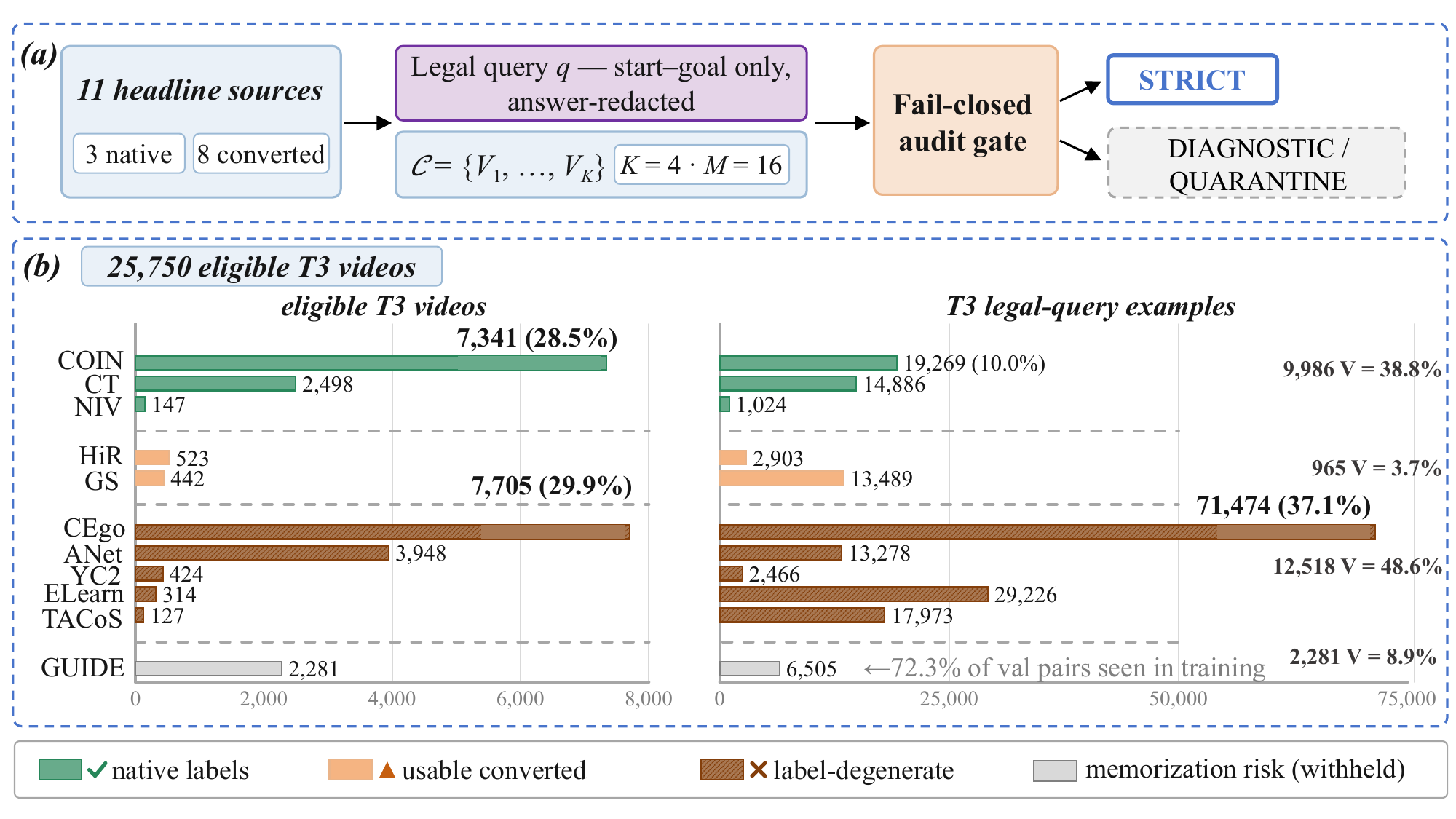}
\captionof{figure}{\textbf{Building the CVSPP benchmark.} (a) Native queries pass lexical screening. (b) Coverage of all eleven sources, including diagnostic subsets. Colors mark label quality and the GUIDE overlap warning. The supplement defines these audit categories.}
\label{fig:benchmark}
\end{minipage}\hfill
\begin{minipage}[t]{.38\textwidth}
\vspace{0pt}\centering
{\fontsize{8}{9.2}\selectfont\setlength{\tabcolsep}{1.3pt}
\begin{tabular}{@{}llrrrr@{}}\toprule
Source & Labels & \multicolumn{2}{c}{3 actions} & \multicolumn{2}{c}{4 actions}\\
 & & Videos & Examples & Videos & Examples\\\midrule
NIV~\cite{niv2018} & N & 147 & 1,024 & 146 & 877\\
COIN~\cite{coin2019} & N & 7,341 & 19,269 & 4,647 & 11,928\\
CT~\cite{crosstask2019} & N & 2,498 & 14,886 & 2,297 & 12,388\\
YC2~\cite{youcook2_2018} & W & 424 & 2,466 & 398 & 2,016\\
TACoS~\cite{tacos2013} & W & 127 & 17,973 & 127 & 17,846\\
HiR~\cite{hirest2023} & W & 523 & 2,903 & 443 & 2,380\\
GS~\cite{ego4dgoalstep2023} & W & 442 & 13,489 & 393 & 12,521\\
ANet~\cite{anetcaptions2017} & W & 3,948 & 13,278 & 2,057 & 7,439\\
CEgo~\cite{charadesego2018} & R & 7,705 & 71,474 & 7,492 & 63,556\\
GUIDE~\cite{guide2024} & R & 2,281 & 6,505 & 1,213 & 3,156\\
ELearn~\cite{egoexolearn2024} & W & 314 & 29,226 & 314 & 28,912\\
\midrule
EEL~\cite{egoexolearn2024} & T & 177 & 26,018 & 177 & 25,841\\
\bottomrule\end{tabular}}
\captionof{table}{\textbf{CVSPP source inventory.} Training and validation counts. The eleven sources total 25,750 videos and 192,493 three-action examples; EEL is a separate diagnostic. Label types: N native, W weak, R released, T taxonomy. Evaluated plan lengths are defined in Section~\ref{sec:benchmark}.}
\label{tab:benchmark-inventory-r13}
\end{minipage}
\end{figure*}

The benchmark turns existing video annotations into requests for a procedure and its recorded support.
Each example contains a start--goal query, candidate videos, a plan length, and annotated video--interval--plan targets.
Figure~\ref{fig:benchmark} shows the construction and source composition; Table~\ref{tab:benchmark-inventory-r13} gives the inventory.

\paragraph{Sources and task coverage.}
NIV, COIN, and CrossTask (CT) supply native action labels~\cite{niv2018,coin2019,crosstask2019}.
Eight additional sources broaden the construction: YouCook2 (YC2), TACoS, HiREST (HiR), Ego4D GoalStep (GS), ActivityNet Captions (ANet), Charades-Ego (CEgo), GUIDE, and the EgoLearn split of EgoExoLearn (ELearn)~\cite{youcook2_2018,tacos2013,hirest2023,ego4dgoalstep2023,anetcaptions2017,charadesego2018,guide2024,egoexolearn2024}.
The benchmark studies three- and four-action plans on the native sources and three-action plans on the eight converted sources: fourteen settings in total.
Six converted sources derive deterministic action labels from leading tokens in step descriptions; CEgo and GUIDE retain supplied class or step labels.
The supplementary material documents conversion rules, source splits, and the additional taxonomy-based EgoExoLearn (EEL) diagnostic.

\paragraph{Queries and linked references.}
For $T$ annotation records, we form the query from the task and endpoint descriptions and use the action labels as targets.
The reference interval spans the native annotation hull or the converted window's first start to last end.
The query omits the target video index, timestamps, and intermediate states.
Native-source lexical checks remove specified action-name phrases or exclude flagged records; converted-source templates are documented separately.
Examples matching the complete query string and requested length form a finite set of alternative references.
Each alternative retains its own video, interval, and full label sequence; evaluation never mixes parts of different examples.

\paragraph{Candidate regimes.}
Distinct-task candidates test discrimination across tasks; same-task candidates test which demonstration and stage support the requested transition; mixed candidates include both.
Our main studies use $K=4$ videos.
The current NIV, CrossTask, and COIN comparisons use mixed pools: the construction video, two other videos of the same task, and one of a different task. Alternative annotated references are collected within that fixed pool.
The construction metadata distinguishes wrong-video (WV) decoys from the same task, wrong windows (WO) in the reference video, and distinct-task videos (WA). These roles describe designed alternatives, not a guarantee that every unannotated support is invalid.
Exact candidate and reference rules for each study appear in the supplement.

% Original Figure 2 composition with corrected callout; independent figure/table counters.

\section{Task Formulation}
\label{sec:v6contract}
\paragraph{Prediction target.}
Let $q$ describe the start and goal, $\mathcal C=\{V_1,\ldots,V_K\}$ be the candidates, and $T$ the requested number of actions.
The model predicts $\hat y=(\hat k,\hat I,\hat A)$: a video index, an interval in that video, and labels $\hat A=(\hat a_1,\ldots,\hat a_T)$ from the source vocabulary.
A \emph{complete plan} contains these $T$ target labels in source annotation order; it need not include every action inside the interval.
The model observes the query and candidate features, not the target triple.

\paragraph{Joint evaluation.}
Let $\mathcal R(q,\mathcal C,T)$ denote the annotated reference triples. Joint success requires:
\begin{equation}
\begin{split}
J(\hat y)=\mathbf1\{\exists(k,I,A)\in\mathcal R:\;&\hat k=k,\quad\hat A=A,\\[-2pt]
&\operatorname{IoU}(\hat I,I)\ge .5\}.
\end{split}\label{eq:v6joint}
\end{equation}
We also report video accuracy (V), video--interval success (VI), complete-plan accuracy (P), and video--plan success (VP).
VI and VP each require a shared reference for their two components. A plan may match one demonstration and the selected interval another: P and VI then pass, but J fails. Rates use all declared cases; invalid predictions count as failures and malformed references invalidate evaluation.
The earlier FULL-SR metric corresponds to VP and omits interval overlap; Table~\ref{tab:benchmark-results-r13} separates this adaptation inventory from the full joint evaluation.
Primary intervals use a 16-bin normalized annotation grid. Section~\ref{sec:reference-sensitivity} examines how changing this reference representation affects the result.

\section{Method: One-Step Evidence Fusion}
\label{sec:v6method}
\begin{figure*}[t]
\centering
\includegraphics[width=\textwidth]{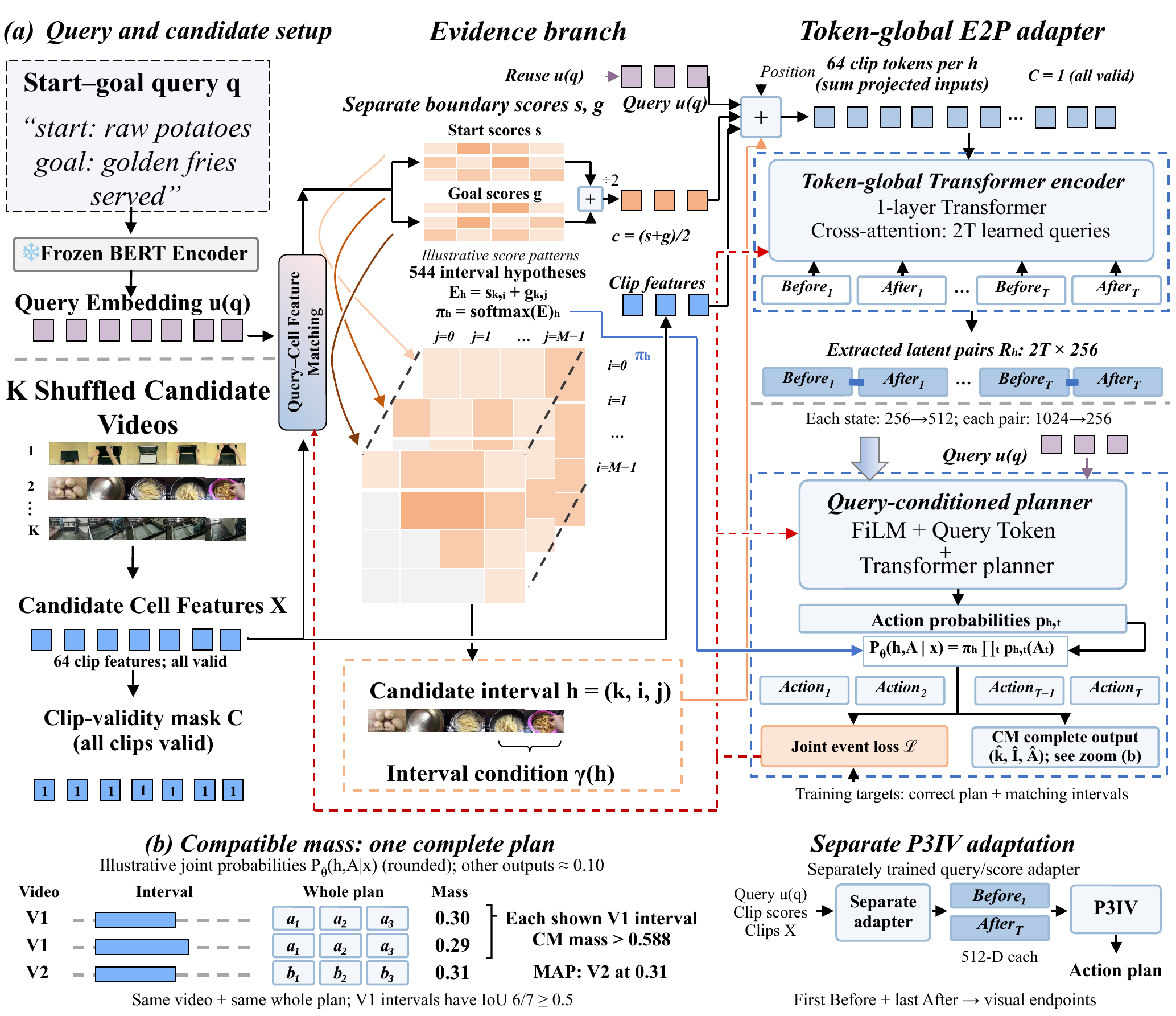}
\caption{\textbf{OSEF connects evidence to complete plans.} (a) OSEF scores intervals from their boundaries and plans over all clips. Red dashes show training gradients. (b) Two shared-start intervals support one complete plan. The \protect\hyperlink{osef-cm-examples}{full-bank examples} give the decoded intervals. A separate P3IV adapter~\cite{p3iv2022} supplies visual endpoints. The main path depicts NIV; heatmaps and probabilities are illustrative.}
\label{fig:framework}\label{fig:manual-global-024}
\end{figure*}

\paragraph{Design principle.}
OSEF delays the final evidence choice until planning is complete.
It scores boundaries across candidates, forms interval-conditioned state representations, and selects a linked output.
``One-step'' refers to this delayed commitment, not a single network layer or forward pass.
Figure~\ref{fig:framework} connects the evidence branch, token-global adapter, and query-conditioned planner.

\subsection{Boundary Scoring}
Each candidate video is represented by $M=16$ ordered clip features $X_{km}$. A validity mask $C_{km}$ marks observed clips; the global profile here requires all 16 clips to be valid. We use \emph{cell} only for a temporal clip, not for a benchmark setting.
With $K=4$ videos, an interval hypothesis $h=(k,i,j)$ selects a video and first/last clip indices, $0\le i\le j<M$.
Its interval is $I_h=[i/M,(j+1)/M)$.
There are $H=KM(M+1)/2=544$ such hypotheses.
We first encode the full start--goal query with frozen BERT~\cite{bert2019} and average its valid token embeddings to obtain $u(q)\in\mathbb R^{768}$.
This vector represents query content; its coordinates are features, not action labels or probabilities.
Separate learned projections map $u(q)$ to start and goal representations and map each clip to the same feature space.
Writing their normalized vectors as $\bar s_q$, $\bar g_q$, and $\bar v_{km}$, we compute scaled cosine similarities:
\begin{equation}
s_{km}=10\bar v_{km}^{\top}\bar s_q,\qquad
g_{km}=10\bar v_{km}^{\top}\bar g_q.
\label{eq:boundary-similarities-r12}
\end{equation}
These scores measure learned agreement with the requested endpoints.
An interval combines its start and goal scores, then competes with all other intervals:
\begin{equation}
E_h=s_{ki}+g_{kj},\qquad
\pi_h=\frac{\exp E_h}{\sum_{h'}\exp E_{h'}}.
\label{eq:p23-boundary-token}
\end{equation}
Separate boundary scores allow the start and goal to occur in different clips, avoiding the singleton preference of averaging one shared score over an interval.
The softmax converts signed scores into nonnegative weights summing to one; $\pi_h$ expresses relative model preference over intervals.

\subsection{Interval-Conditioned Planning}
For every hypothesis $h$, the global planner retains all $KM=64$ clip features.
Each token combines its visual feature, query embedding, position, scalar score $c_{km}=(s_{km}+g_{km})/2$, and geometry relative to $h$.
Geometry $\gamma(h)$ indicates whether the clip belongs to the hypothesized video, its offsets from the two boundaries, and the interval length.
It lets the same video collection support different conditional plans.
The \emph{evidence-to-plan adapter} contextualizes these tokens with a Transformer~\cite{vaswani2017}.
Its $2T$ learned queries extract $T$ pairs of latent state representations, $R_h\in\mathbb R^{2T\times256}$, one pair per action position. Before/After denote learned latent roles, without separate state supervision.
Each latent is projected to 512 dimensions; the concatenated pair is then mapped to one 256-dimensional step token.
The query scales and shifts each step through FiLM~\cite{film2018}; a query token then joins the steps in the Transformer planner, which produces action logits $Z_\theta(x,h)$.
Row $t$ is normalized over action classes as $p_{h,t}(a)=\operatorname{softmax}(Z_\theta(x,h)_{t,:})_a$.

The resulting distribution is over complete outputs:
\begin{equation}
P_\theta(h,A\mid x)=\pi_h\prod_{t=1}^{T}p_{h,t}(A_t),
\label{eq:joint-model}
\end{equation}
Here $x$ denotes the query and candidate features, and $\theta$ the learned parameters.
Action probabilities factorize conditional on $h$ and $x$, although planner self-attention couples their representations.
Each term therefore links one interval to an entire plan.
These normalized model probabilities need not be empirically calibrated.

\begin{samepage}
For training example $(k^*,I^*,A^*)$, we minimize the negative log probability of the correct plan across acceptable intervals:
\begin{equation}
\mathcal L=-\log\!\!\sum_{\substack{h:k_h=k^*\\\operatorname{IoU}(I_h,I^*)\ge.5}}
\pi_h\prod_{t=1}^{T}p_{h,t}(A_t^*).
\label{eq:p23-canonical-event}
\end{equation}
\end{samepage}
Training uses the original construction example; evaluation may accept additional annotated examples.
The loss increases probability for the annotated plan across intervals supporting that example; no separate state-supervision loss is used.
The primary NIV results use this global model.
COIN averages representations extracted from all clips and from clips inside each hypothesis.
The primary CrossTask comparison instead uses an adapted locator and our window planner (Table~\ref{tab:benchmark-results-r13}(b)); full implementations appear in the supplement.

The model evaluates 544 interval conditions, each with clip inputs, rather than treating intervals as 544 input tokens.
Chunking reduces working memory; computational cost is analyzed in the supplement.

\begin{table*}[t]
\centering
{\small\bfseries (a) VSPP-to-CVSPP benchmark: video--plan success (VP, \%)}\par\vspace{2pt}
{\fontsize{8}{9.2}\selectfont\setlength{\tabcolsep}{1.5pt}
\begin{tabular*}{\textwidth}{@{\extracolsep{\fill}}l*{14}{r}@{}}\toprule
& \multicolumn{2}{c}{CrossTask~\cite{crosstask2019}} & \multicolumn{2}{c}{COIN~\cite{coin2019}} & \multicolumn{2}{c}{NIV~\cite{niv2018}} & \multicolumn{8}{c}{Converted sources (diagnostic)} \\
System & $T{=}3$ & $T{=}4$ & $T{=}3$ & $T{=}4$ & $T{=}3$ & $T{=}4$ & YC2~\cite{youcook2_2018} & TACoS~\cite{tacos2013} & HiR~\cite{hirest2023} & GS~\cite{ego4dgoalstep2023} & ANet~\cite{anetcaptions2017} & CEgo~\cite{charadesego2018} & GUIDE$^\ddagger$~\cite{guide2024} & ELearn~\cite{egoexolearn2024} \\\midrule
Majority floor & 0.44 & 0.47 & 0.45 & 0.69 & 4.50 & 4.17 & 8.54 & 5.47 & 6.63 & 0.68 & 1.99 & 0.07 & 0.70 & 0.86 \\
ViterbiPlanNet$^*$~\cite{viterbiplannet2026} & 8.05 & 4.51 & 15.90 & 12.77 & 14.71 & 12.15 & 8.35 & 5.47 & 7.16 & 0.89 & 1.34 & 0.11 & 7.05 & 0.86 \\
MTID$^*$~\cite{mtid2025} & 3.24 & 1.13 & 2.96 & 1.24 & 6.91 & 4.17 & 8.54 & 5.47 & 6.63 & 0.68 & 1.99 & 0.02 & 0.00 & 0.86 \\
SCHEMA$^*$~\cite{schema2024} & 7.54 & 3.91 & 16.28 & 13.53 & 13.51 & 10.07 & 8.54 & 5.47 & 6.63 & 0.72 & 2.11 & 0.11 & 0.31 & 0.86 \\
P3IV$^*$~\cite{p3iv2022} & 7.05 & 4.03 & 7.79 & 4.63 & 12.91 & 10.42 & 8.35 & 5.47 & 7.16 & 0.25 & 0.50 & 0.08 & 3.49 & 0.86 \\
KEPP$^*$~\cite{kepp2024} & 4.54 & 2.02 & 0.00 & 0.45 & 9.61 & 10.42 & 3.80 & 5.47 & 6.63 & 0.64 & 1.99 & 0.03 & 0.23 & 0.86 \\
PDPP$^*$~\cite{pdpp2023} & 2.22 & 1.78 & 12.09 & 9.56 & 6.61 & 8.68 & 5.12 & 5.47 & 2.39 & 0.75 & 1.99 & 0.03 & 1.39 & 0.86 \\
ActionDiffusion$^*$~\cite{actiondiffusion2025} & 4.40 & 1.11 & 14.21 & 9.80 & 10.21 & 10.07 & 4.17 & 0.98 & 0.80 & 0.14 & 1.99 & 0.01 & 2.40 & 0.26 \\
PlanLLM$^*$~\cite{planllm2025} & 8.04 & 3.62 & 17.11 & 13.19 & 14.71 & 7.29 & 7.40 & 4.90 & 6.37 & 0.21 & 1.53 & 0.08 & 7.44 & 0.86 \\
Skip-Plan$^*$~\cite{skipplan2023} & 1.08 & 0.83 & 0.04 & 0.14 & 4.50 & 4.17 & 8.54 & 5.47 & 6.63 & 0.36 & 1.99 & 0.00 & 0.00 & 0.86 \\
Shared-plan OSEF & \textbf{58.09} & \textbf{35.39} & \textbf{67.30} & \textbf{46.19} & \textbf{66.07} & \textbf{48.26} & 8.54 & 5.47 & 7.16 & 1.25 & 2.30 & 0.01 & 41.44 & 0.86 \\
\bottomrule\end{tabular*}}\par
\vspace{2pt}
{\small\bfseries (b) Joint success and its components on current development cohorts (\%)}\par\vspace{2pt}
{\footnotesize\setlength{\tabcolsep}{2.0pt}
\begin{tabular*}{\textwidth}{@{\extracolsep{\fill}}lcccrrrrrr@{}}\toprule
Dataset/system & $T$ & $N$ & Decision & $V$ & $VI_{.5}$ & $P$ & $VP$ & $J_{.5}$ & $\Delta J$\\\midrule
NIV~\cite{niv2018}: OSEF & 3 & 333 & MAP & $55.76\!\pm\!3.29$ & $34.93\!\pm\!1.14$ & $70.27\!\pm\!1.08$ & $40.44\!\pm\!1.65$ & $26.93\!\pm\!0.76$ & $0$\\
NIV~\cite{niv2018}: OSEF & 3 & 333 & CM & $56.66\!\pm\!2.29$ & $38.34\!\pm\!1.65$ & $70.27\!\pm\!1.08$ & $41.24\!\pm\!0.76$ & $\mathbf{29.33\!\pm\!1.21}$ & $+2.40$\\
\addlinespace[2pt]
NIV~\cite{niv2018}: OSEF & 4 & 288 & MAP & $49.19\!\pm\!1.75$ & $28.12\!\pm\!2.43$ & $52.43\!\pm\!2.17$ & $29.98\!\pm\!2.36$ & $18.29\!\pm\!3.48$ & $0$\\
NIV~\cite{niv2018}: OSEF & 4 & 288 & CM & $50.35\!\pm\!0.92$ & $33.45\!\pm\!2.21$ & $52.43\!\pm\!2.17$ & $30.44\!\pm\!1.45$ & $\mathbf{21.06\!\pm\!1.57}$ & $+2.78$\\
\addlinespace[2pt]
CrossTask~\cite{crosstask2019}: locator$^\dagger$ & 3 & 2,028 & MAP & $56.67\!\pm\!0.20$ & $38.31\!\pm\!0.88$ & $63.08\!\pm\!1.32$ & $34.57\!\pm\!0.77$ & $24.62\!\pm\!0.35$ & $0$\\
CrossTask~\cite{crosstask2019}: locator$^\dagger$ & 3 & 2,028 & CM & $56.99\!\pm\!0.23$ & $41.45\!\pm\!0.97$ & $63.63\!\pm\!1.18$ & $35.12\!\pm\!1.03$ & $\mathbf{27.25\!\pm\!0.72}$ & $+2.63$\\
\addlinespace[2pt]
COIN~\cite{coin2019}: OSEF & 3 & 4,685 & MAP & $61.01\!\pm\!0.65$ & $33.74\!\pm\!0.31$ & $69.90\!\pm\!0.29$ & $43.49\!\pm\!0.86$ & $24.77\!\pm\!0.66$ & $0$\\
COIN~\cite{coin2019}: OSEF & 3 & 4,685 & CM & $61.17\!\pm\!0.19$ & $42.78\!\pm\!0.61$ & $70.59\!\pm\!0.32$ & $44.30\!\pm\!0.33$ & $\mathbf{31.57\!\pm\!0.28}$ & $+6.79$\\
\bottomrule\end{tabular*}}
\caption{\textbf{Benchmark breadth and joint success.} (a) Historical VP. The six native settings use distinct-task pools. The eight converted settings are diagnostic. $^*$Adapted planners use frozen OSEF evidence and retain their original validation-selected schedules. $^\ddagger$GUIDE is excluded from ranking. (b) Mixed $K=4$ development cohorts, three-seed means $\pm$ sample SD. Only decoding changes within each pair. V/VI/P/VP denote video, video--interval, plan, and video--plan success. J requires all three outputs to match one reference at IoU $\ge .5$; $\Delta J$ is the paired gain before rounding (percentage points). Bold marks the best native VP or paired J. $^\dagger$CONQUER-derived locator~\cite{conquer2021} plus planner. Protocols and numerical controls appear in the supplement.}
\label{tab:benchmark-results-r13}
\end{table*}

\subsection{Compatible-Mass Decoding}
Maximum a posteriori (MAP) decoding selects the single most probable $(h,A)$ under Eq.~\ref{eq:joint-model}.
The evaluation, however, accepts an interval that sufficiently overlaps its reference.
Several nearby hypotheses can therefore support the same decision.
Illustratively, compatible outputs of probability $0.30$ and $0.29$ can together outweigh an incompatible output of probability $0.31$.

\begin{samepage}
For a proposed output $(d,A)$, where $d=(k_d,i_d,j_d)$ is an interval hypothesis and $I_d$ its interval, define the compatible mass
\begin{equation}
U(d,A)=\!\!\sum_{\substack{h:k_h=k_d\\\operatorname{IoU}(I_h,I_d)\ge.5}}
\pi_h\prod_{t=1}^{T}p_{h,t}(A_t).
\label{eq:r14-compatible-mass}
\end{equation}
\end{samepage}
The sum preserves the same video and the same complete plan while allowing overlapping intervals.
Each interval contributes its probability of that plan, even when the plan is not its highest-probability sequence.
In particular, it sums products of action probabilities within each hypothesis; it does not combine separate action positions across hypotheses.
Compatibility is tested against each proposed interval, so overlapping sets need not form disjoint groups.

We search all intervals and the deduplicated pool of their most probable plans, including the retained baseline plan.
Among maximizers, prefer the baseline if present, then follow the fixed order.
This minimizes model-expected rejection loss within the pool.
This does not guarantee higher empirical joint success or optimality over every possible plan.
A full-action search reproduces the pool output on all 13,536 evaluated NIV and CrossTask model--case pairs; COIN uses the stated pool. The supplement gives the proof, complete search checks, and numerical controls.

\paragraph{VSPP-to-CVSPP adaptation.}
A VSPP planner expects visual states; CVSPP supplies a query and candidate recordings.
Our shared retrieval-and-state interface provides predicted endpoint features or a state sequence, and we train the planning component on CVSPP targets.
The front end is fixed when comparing inherited planners.
In Table~\ref{tab:benchmark-results-r13}(a), shared-plan OSEF predicts one plan from global candidate features. The interval-conditioned formulation instead lets different hypotheses support different plans. Table~\ref{tab:components-r13}(c) separately examines the visual-state interface supplied to P3IV.
Donor-specific transformations are detailed in the supplement.

\section{Experiments}
\label{sec:exp}
We evaluate transfer, joint correctness, component effects, and temporal-reference sensitivity.

\subsection{Experimental Setup}
\paragraph{Benchmark and comparisons.}
Table~\ref{tab:benchmark-results-r13}(a) reports historical VP in fourteen settings. The six native settings use distinct-task pools.
Panel (b) evaluates current video--interval--plan decisions with all five axes.
Within each panel-(b) pair, the weights, candidate records, interval-conditioned predictions, and references remain fixed; only decoding changes.
The NIV global model uses 333/288 development cases at $T=3/4$; COIN uses a local/global mixture on 4,685 cases.
CrossTask combines a CONQUER-derived locator~\cite{conquer2021} with our window planner on 2,028 cases, evaluating every interval before the complete-output decision.
This is a study within each system, not a ranking across architectures.

\paragraph{Implementation and reporting.}
The current NIV controls use frozen InternVideo-Next features~\cite{wang2026internvideonext} and BERT query embeddings~\cite{bert2019}; the supplement identifies all feature releases, projections, and model configurations.
Current NIV input conditions share initialization and training budgets, with the primary checkpoint fixed at epoch 100.
We report three-seed means and sample standard deviations, and compute paired differences before rounding.
Historical adaptations retain their original validation-selected schedules.
The reported development records have been reused during research; all cohort definitions, seed results, and supplementary diagnostics are provided separately.

\subsection{RQ1: Do VSPP Planners Transfer to CVSPP?}
The broad matrix in Table~\ref{tab:benchmark-results-r13}(a) makes the transfer problem visible across both planning methods and data sources.
The nine adaptations receive predicted support through a shared interface, while the majority row supplies a constant-sequence floor under that same evidence.
On the six native settings, shared-plan OSEF has the highest VP in this inventory.
The gap shows the value of studying evidence access alongside planning, but VP alone cannot establish complete CVSPP success.

Converted sources expose a different issue: several methods approach the majority floor when weak labels collapse distinct descriptions into the same sequence.
GUIDE's $41.44\%$ remains diagnostic under the original overlap flag; repeated query--plan pairs motivate further checks but do not alone establish memorization. On CEgo, OSEF's $0.01\%$ is below the $0.07\%$ floor.
These outcomes are retained because benchmark coverage includes cases where a construction or representation gives little useful discrimination.
A further 240-fit study crosses three evidence interfaces, two prediction heads, and five seeds on the eight converted sources. Every condition mean remains below $2.5\%$ joint success, exposing a demanding cross-source transfer problem. The supplement reports the construction, label analyses, and complete results.

% Candidate A: independent single-column float. Requires caption + booktabs.
\begin{table}[t]
\centering
{\fontsize{9}{10.4}\selectfont\setlength{\tabcolsep}{2pt}\renewcommand{\arraystretch}{1.08}
\begin{tabular*}{\linewidth}{@{}l@{\hspace{7pt}}l@{\extracolsep{\fill}}rr@{}}\toprule
$T$ & Input & $P$ & $J_{.5}$\\
\midrule
3 & Query only & $71.87\pm1.21$ & $8.01\pm2.27$\\
 & Query and video & $70.27\pm1.08$ & $26.93\pm0.76$\\
 & Oracle video$^\dagger$ & $70.17\pm1.21$ & $31.63\pm2.13$\\
 & Oracle window$^\dagger$ & $70.07\pm1.35$ & $64.36\pm1.35$\\
\midrule
4 & Query only & $54.05\pm0.40$ & $4.40\pm3.56$\\
 & Query and video & $52.43\pm2.17$ & $18.29\pm3.48$\\
 & Oracle video$^\dagger$ & $52.66\pm2.83$ & $27.08\pm2.08$\\
 & Oracle window$^\dagger$ & $52.43\pm2.76$ & $48.96\pm2.28$\\
\bottomrule\end{tabular*}}
\captionsetup{labelsep=none}
\caption{(a). \textbf{Input and oracle evidence.} NIV~\cite{niv2018}, $T=3/4$, $N=333/288$, joint MAP for the query rows. P: complete-plan accuracy; J: joint success. All panels report percentages (three-seed mean $\pm$ sample SD). Trained input controls match initialization and budgets. $\dagger$Oracle rows use the supplied construction video or window to select an interval, then decode its plan.}
\label{tab:components-r13}\label{tab:input-control}
\end{table}

% Candidate A: independent single-column float. Requires caption + booktabs.
\begin{table}[t]
\ContinuedFloat
\centering
{\fontsize{9}{10.4}\selectfont\setlength{\tabcolsep}{2pt}\renewcommand{\arraystretch}{1.08}
\begin{tabular*}{\linewidth}{@{}l@{\hspace{7pt}}l@{\extracolsep{\fill}}rr@{}}\toprule
$T$ & Context / scalar score & MAP & CM\\
\midrule
3 & Global / with score & $26.93\pm0.76$ & $29.33\pm1.21$\\
 & Global / no score & $27.23\pm1.42$ & $30.43\pm0.97$\\
 & Interval / with score & $25.53\pm1.88$ & $28.83\pm1.88$\\
\midrule
4 & Global / with score & $18.29\pm3.48$ & $21.06\pm1.57$\\
 & Global / no score & $18.98\pm1.60$ & $19.68\pm1.00$\\
 & Interval / with score & $18.63\pm1.64$ & $21.06\pm0.72$\\
\bottomrule\end{tabular*}}
\captionsetup{labelsep=none}
\caption{(b). \textbf{Context and score controls.} NIV~\cite{niv2018}, joint success ($T=3/4$, $N=333/288$). MAP and CM reuse the same model outputs. No score zeros the scalar feature; interval access still searches all 544 hypotheses.}
\label{tab:core-contribution}\label{tab:core-contribution-047}
\end{table}

% Candidate A: independent single-column float. Requires caption + booktabs.
\begin{table}[t]
\ContinuedFloat
\centering
{\fontsize{9}{10.4}\selectfont\setlength{\tabcolsep}{2pt}\renewcommand{\arraystretch}{1.08}
\begin{tabular*}{\linewidth}{@{\extracolsep{\fill}}lrr@{}}\toprule
Planner input & $P$ & $J_{.5}$\\
\midrule
Predicted endpoints & $11.21\pm1.35$ & $7.71\pm0.87$\\
All clips + scores & $12.71\pm5.90$ & $7.31\pm2.11$\\
All clips, no scores & $5.31\pm1.39$ & $4.40\pm0.17$\\
Endpoint-only tokens & $15.42\pm1.81$ & $8.21\pm0.46$\\
All clips, shuffled scores & $10.81\pm5.56$ & $7.01\pm2.21$\\
\bottomrule\end{tabular*}}
\captionsetup{labelsep=none}
\caption{(c). \textbf{Transferring to P3IV.} Separate NIV~\cite{niv2018} construction for P3IV~\cite{p3iv2022}, $T=3$, $N=333$; fixed retriever ($V=59.16$, $VI_{.5}=44.14$), Viterbi decoding. Endpoint-only tokens mask other clips. Token adapters also add query access and parameters. The supplement defines all input transformations.}
\label{tab:adapter-transfer}
\end{table}

\subsection{RQ2: Are Plans Correctly Supported?}
The complete-output evaluation in Table~\ref{tab:benchmark-results-r13}(b) measures what the new task requires.
On NIV with three actions, MAP obtains $70.27\%$ complete-plan accuracy but only $26.93\%$ joint success.
Compatible mass preserves that aggregate plan accuracy while increasing video--interval success from $34.93\%$ to $38.34\%$ and joint success to $29.33\%$.
Thus a better final decision can improve the connection between a plan and its evidence without increasing the number of correct plans.

Compatible mass improves mean joint success in all four primary pairs.
The NIV gains are $2.40$ and $2.78$ percentage points for three and four actions.
CrossTask and COIN also improve without retraining or additional observations. The supplement gives per-seed comparisons and numerical controls.

\begin{figure}[t]
\centering
\includegraphics[width=\columnwidth]{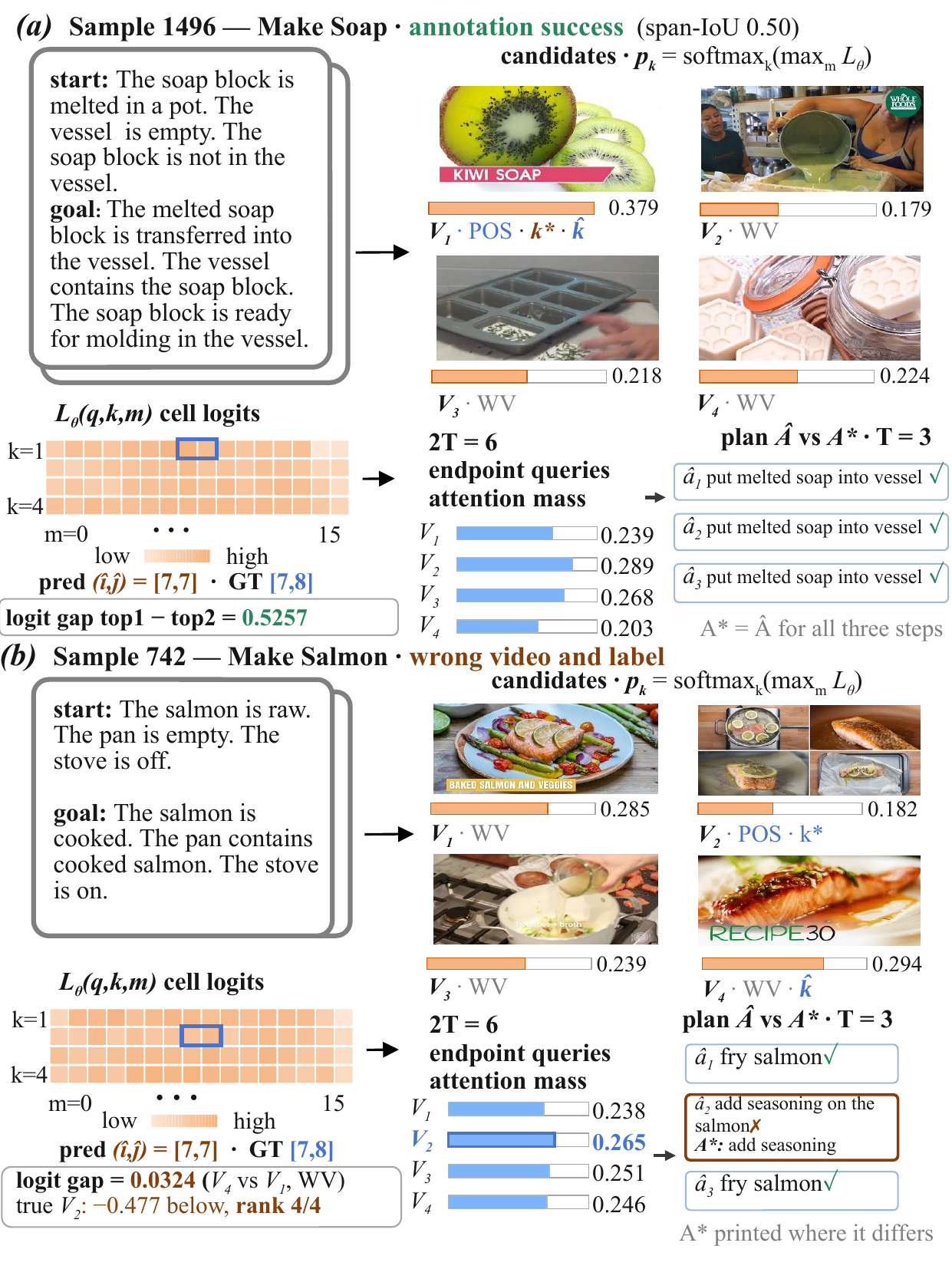}
\caption{\textbf{When evidence supports or misleads a plan.} Saved same-task COIN hard-selection examples: (a) correct video and action labels; (b) wrong video and a different second label. The case analysis and supplement explain the numerical details.}
\label{fig:case}
\end{figure}

\subsection{RQ3: Which Components Help?}
Table~\ref{tab:components-r13} separates input evidence, token access, scalar scores, and the interface presented to a donor planner.
On three-action NIV, the query-only fit has higher plan accuracy ($71.87\%$ vs.\ $70.27\%$) but lower joint success ($8.01\%$ vs.\ $26.93\%$) than the query-and-video fit.
The query can predict many labels, but that does not establish the supporting recording or interval.
The privileged controls provide a complementary diagnostic. Supplying the reference video raises NIV joint success to $31.63/27.08\%$ at three/four actions; supplying its window raises it to $64.36/48.96\%$. This substantial remaining gap motivates better evidence localization, while oracle access itself is unavailable at inference. Matched MAP-decoded CrossTask and COIN controls likewise show a larger visual-input benefit for joint success than for plan accuracy; full results and query-stratified oracle diagnostics appear in the supplement.

Compatible mass improves mean joint success in all six NIV access/score conditions.
The controls do not establish a universal advantage for every part of the global model: at three actions, the unscored global fit reaches $30.43\%$ joint success with CM, compared with $29.33\%$ for the scored fit.
With clip scores and CM, global and interval-only access both reach $21.06\%$ joint success at four actions.
This separates output selection from the representation feeding it. A separate fixed-locator study likewise finds that broader clip access can improve plan accuracy without improving joint success; its paired controls are reported in the supplement.

For P3IV, a fixed retriever keeps video and video--interval success at $59.16\%$ and $44.14\%$.
Applying the token adapter only to its predicted boundary features and scores gives $8.21\%$ joint success, compared with $7.71\%$ for raw endpoint features and $7.31\%$ for all-clip tokens with scores.
The result identifies an important adaptation choice: preserving more candidate information is useful to test, but need not help every donor interface.

\subsection{RQ4: Temporal-Reference Sensitivity}
\label{sec:reference-sensitivity}
CVSPP introduces a temporal-reference question alongside procedure prediction.
Keeping every NIV three-action prediction and reference-example identity fixed, we replace the 16-bin reference projection with the unsnapped annotation endpoints.
MAP/CM joint success changes from $26.93/29.33\%$ to $19.42/16.32\%$: the gain reverses from $+2.40$ to $-3.10$ points.
The plan, candidate set, and predicted interval do not change.
This controlled reversal identifies reference representation as part of the evaluation problem, not merely a plotting choice. Expanding the final interval search also fails to remove this reversal in the checked three-action systems (supplement).

The supplement extends this analysis across three reference conventions, five IoU thresholds, and all fourteen retained model--seed combinations.
It also reports candidate-count, feature-perturbation, oracle-evidence, and source-conversion studies.
These analyses identify further challenges for supported planning.

\subsection{Case Analysis}
Figure~\ref{fig:case} shows two recorded COIN hard-selection cases. Both predict bin $[7,7]$ against reference $[7,8]$: their bin-index overlap is $0.50$, but only the soap output also matches the reference video and action labels. The salmon output selects $V_4$ instead of $V_2$ and differs at the second action label: a plausible plan cannot repair a wrong evidence choice. These baseline cases do not demonstrate a CM gain.

\section{Conclusion}
CVSPP associates a predicted procedure with its supporting video interval. Our eleven-source benchmark and nine planner adaptations establish a basis for studying this task. OSEF connects boundary scoring, interval-conditioned planning, and compatible-mass decoding. The experiments distinguish correct plans from correctly supported plans, characterize transfer and evidence-access challenges, and reveal temporal-reference sensitivity.

\paragraph{Limitations.}
Results use repeatedly studied development data, not a new independent test set.
Deterministic queries, lexical overlap, and weak labels can expose dataset regularities, and finite references do not enumerate every semantically valid solution.
NIV target order follows annotation lists and may differ from chronology; normalized reference coordinates are not independently verified physical timestamps.
Adaptation protocols differ in input access and capacity, and CM gains depend on reference geometry.
The supplementary material gives the detailed definitions, negative results, mathematical qualifications, and remaining validation questions.

\FloatBarrier
\typeout{DRAFT_MAIN_CONTENT_END_PAGE: \thepage}
\clearpage
\twocolumn
\raggedbottom
\typeout{LLM_USAGE_START_PAGE: \thepage}
\section*{LLM Usage Statement}
We used OpenAI Codex and ChatGPT extensively across research and writing: literature discovery and citation checking, research ideation and control design, mathematical and methodological analysis, data/feature-processing and model/evaluation code implementation and debugging, experiment configuration, orchestration and monitoring, numerical analysis and plotting code, and manuscript drafting and revision. We also used the TRAE AI assistant for source-based manuscript and supplementary editing, cross-document references and build checks. Suggestions from these tools are not experimental observations. Reported measurements are obtained from the specified executed programs and retained outputs, with their evaluation conditions stated separately. The authors' original PowerPoint figures were retained; these tools assisted text, notation, connector, and layout refinements in copies. Original video thumbnails were retained. Task and architecture figures are explanatory diagrams; the case figure reports saved baseline predictions. Numerical plots are generated from the stated experiment data. The authors remain responsible for checking and defending the claims, citations, code, results and final scientific decisions.

\typeout{DRAFT_REFERENCES_START_PAGE: \thepage}
{\small
\bibliographystyle{ieeenat_fullname}
\bibliography{refs}
}

\widowpenalty=150
\PlatformSupplementLayout
\onecolumn
\raggedbottom
\appendix
\PlatformSupplementPhase
\begin{center}
{\Large\bfseries\thetitle\par}
\smallskip
{\large Supplementary Material\par}
\end{center}
\renewcommand{\thefigure}{S\arabic{figure}}
\renewcommand{\thetable}{S\arabic{table}}
\renewcommand{\theequation}{S\arabic{equation}}
\section*{Start here: CVSPP in real frames}
CVSPP takes a start--goal query and returns an ordered plan, its video and its time interval. Both clips start with the central carrier alone and end with both chassis sections and their locking pins fitted.
\par\smallskip
\noindent\begin{minipage}{\linewidth}
\centering
\hypertarget{osef-order-example}{}\includegraphics[width=.96\linewidth]{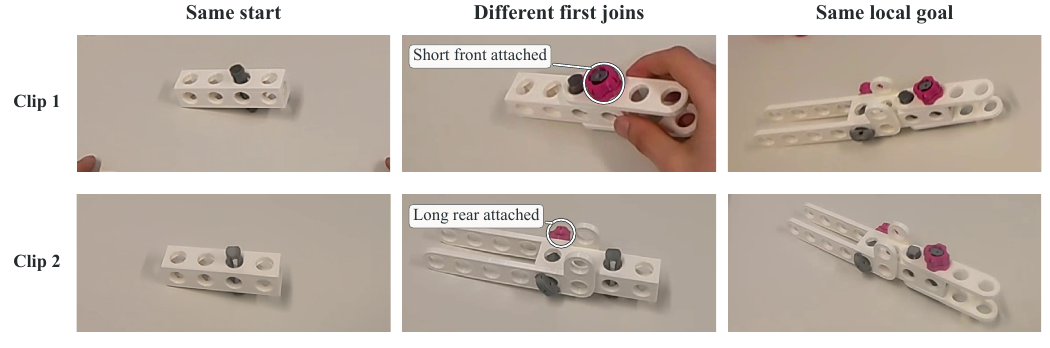}
\par\raggedright\small
\textbf{Figure A.1. One assembly goal, two observed orders.} In the top recording, the short front chassis and its pin are fitted first. In the bottom recording, the long rear chassis and its pin are fitted first. Both local windows end with both groups fitted. Cropped and marked IndustReal v2 TRAIN frames from Schoonbeek et al.~\cite{industreal2024}, Apache~2.0. This two-step illustration is outside the eleven-source evaluation.
\end{minipage}
\par\smallskip

\noindent\textbf{Which outputs pass?}
Let $A$ mean front then rear, and $B$ mean rear then front. For this worked example, the two valid references are $(1,I_a,A)$ and $(2,I_b,B)$. Use the \hyperlink{osef-timing-example}{tool-exchange clip in Fig.~F.1} as a different-task candidate, clip~3, with pictured window $I_c$.
\par\smallskip
\begingroup
\small
\renewcommand{\arraystretch}{1.20}
\begin{tabularx}{\linewidth}{@{}p{.17\linewidth}X p{.19\linewidth}p{.17\linewidth}@{}}
\toprule
Prediction & What the cited video and interval show & Passes & Fails \\
\midrule
$(1,I_a,A)$ & The first clip and interval show front then rear. & $V,VI,P,VP,J$ & --- \\
$(2,I_b,B)$ & The second clip and interval show rear then front. & $V,VI,P,VP,J$ & --- \\
$(1,I_a,B)$ & The interval is right for clip~1. The proposed order is from clip~2. & $V,VI,P$ & $VP,J$ \\
$(1,I_a,C)$ & Plan $C$ repeats the front group and omits the rear group visible at the goal. & $V,VI$ & $P,VP,J$ \\
$(1,I_{\rm late},A)$ & The order is right for clip~1. The late window misses the visible change at the front retainer. & $V,P,VP$ & $VI,J$ \\
$(3,I_c,A)$ & Clip~3 shows a tool exchange. The plan itself matches one valid order. & $P$ & $V,VI,VP,J$ \\
\bottomrule
\end{tabularx}
\par\endgroup
\smallskip
\noindent
$V$ checks the video, $P$ the complete ordered plan, $VI$ the video and interval, $VP$ the video and plan, and $J$ all three together.
Every combined check uses one reference at a time. The table scores illustrative predictions.
\par\smallskip\noindent
\includegraphics[width=\linewidth]{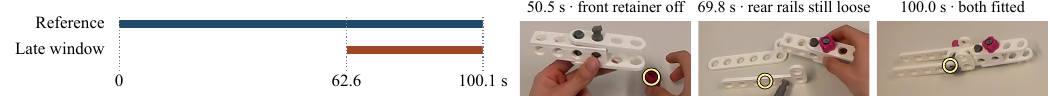}\par
\noindent\textbf{Why the late window fails.}
In clip~1 the front retainer is on by $61.0$~s. The late window covers the last six cells of the $100.1$~s crop, so IoU is $6/16=0.375$. See the \hyperlink{osef-timing-example}{two-view example} for uncertainty near $0.5$.
\par\medskip
\noindent\textbf{A measured order hit and miss.}
Given 49 ordered frames per clip and both answer placements, frozen Qwen3.5-9B~\cite{qwen35_2026} chose front-first in both clips. The middle frames show a hit in clip~1 and a miss in clip~2. This external check tests order alone. See \path{data/NATURAL_ORDER_OPENING_001.json} for outputs.
\par\smallskip
\noindent\textbf{Next.} Removal (\hyperlink{osef-maintenance-example}{Fig.~A.2}), occlusion (\hyperlink{impact-visibility}{Fig.~E.1}) and timing (\hyperlink{osef-timing-example}{Fig.~F.1}, \hyperlink{osef-tray-example}{Fig.~F.2}).
\par\smallskip
\noindent See the corrections to main \hyperref[app:compact-fig3-example]{Fig.~3(b)} and \hyperref[app:compact-fig4-cases]{Fig.~4(a)} and the saved CVSPP outputs in \hyperref[app:compact-fig4-cases]{Sec.~C.1}.

\clearpage
\noindent\textbf{A second task and an ambiguous visual state}
\par\smallskip
\noindent\begin{minipage}{\linewidth}
\centering
\hypertarget{osef-maintenance-example}{}\includegraphics[width=.96\linewidth]{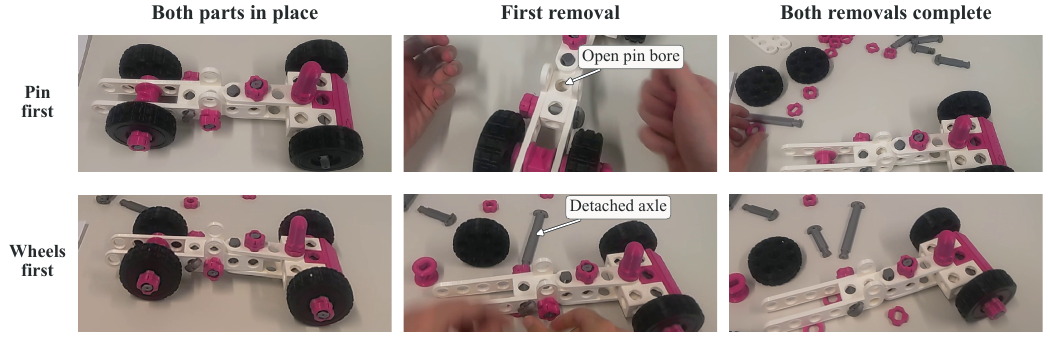}
\par\raggedright\small
\textbf{Figure A.2. The same components can also come apart in different orders.}
Both local windows remove a connector pin and the rear wheel assembly. In the top recording the pin is free while the wheels remain mounted. In the bottom recording the rear assembly is already free while the pin is still in place.
Cropped and marked external IndustReal v2 TRAIN frames from Schoonbeek et al.~\cite{industreal2024}, Apache~2.0.
\end{minipage}
\par\smallskip

\noindent\textbf{What marks a removal?}
A part is visibly free once it clears the joint. In the first recording, the connector pin comes out between the last seated frame at 9.4~s and the first free frame at 9.8~s. The rear assembly clears later. In the second recording the rear assembly clears first, while the connector pin is still seated between the two operations.
These are visible states from the original footage. The time of a publisher's state change can differ from the frame at which separation is visually clear.
\par\smallskip
\begingroup\small
\begin{tabularx}{\linewidth}{@{}l X X@{}}
\toprule
Recording & First visible completion (s) & Second visible completion (s) \\
\midrule
Pin first & Connector pin $(9.4,9.8]$ & Rear wheels and axle $(17.7,24.1]$ \\
Rear assembly first & Rear wheels and axle $(10.6,18.3]$ & Connector pin $(28.4,32.7]$ \\
\bottomrule
\end{tabularx}
\par\endgroup

\par\medskip
\par
\noindent\begin{minipage}{\linewidth}
\centering
\hypertarget{impact-visibility}{}\includegraphics[width=.90\linewidth]{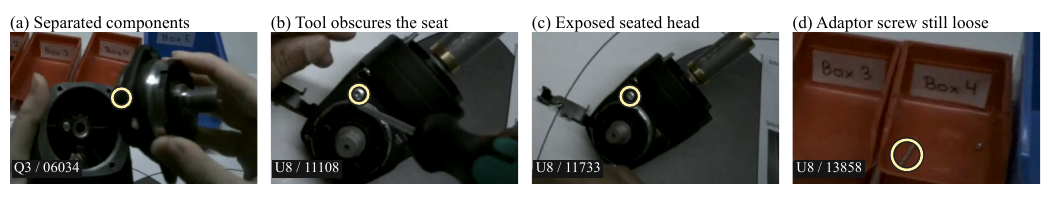}
\par\raggedright\small
\textbf{Figure E.1. Visible states and uncertain completion.} The driver hides the seat in (b). The bearing-screw head is visible in (c), while the exact seating time and torque remain unobserved. The adaptor screw remains loose in (d).
Two external IMPACT v1.1 TRAIN trials illustrate this ambiguity outside the eleven-source evaluation. Cropped and marked frames by Wen et al. (arXiv:2604.10409), \href{https://creativecommons.org/licenses/by-nc-sa/4.0/}{CC BY-NC-SA 4.0}.
\end{minipage}

\noindent\textbf{What remains uncertain?}
Panel (a) is from trial Q3. Panels (b)--(d) are from trial U8. A separate part or a loose screw can show that the assembly is incomplete. A screwdriver over the seat obscures the screw underneath. Once the driver clears, the visible screw head supports a seated-state reading. Those frames bound when the change became observable. Tightening torque remains unmeasured.
\par\medskip
\par\smallskip
\noindent\begin{minipage}{\linewidth}
\centering
\hypertarget{osef-housing-example}{}\includegraphics[width=.96\linewidth]{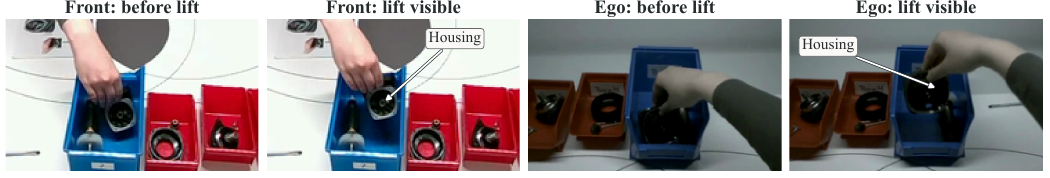}
\par\raggedright\small
\textbf{Figure E.2. A lift seen above the bin rim.}
Both views bracket the rise of the black housing relative to the blue bin. The lift is visible in the later frames.
Cropped and marked external IMPACT v1.1 TRAIN frames by Wen et al. (arXiv:2604.10409), CC BY-NC-SA 4.0.
\end{minipage}
\par\smallskip

\noindent The housing lift has a visible endpoint in both views. On the next page we compare its overlap range with those of two other actions.

\clearpage
\noindent\textbf{When the boundary is uncertain}
\par\smallskip
\noindent\begin{minipage}{\linewidth}
\hypertarget{osef-timing-example}{}\noindent\textbf{The same action can have different visible time bounds.}
In Fig.~F.1, the hand leaves the orange screwdriver and lifts the green one.
For each transition, the bracket spans the last frame visibly before it and the first visibly after it.
The table below compares the midpoint windows and the full ranges of endpoints.
\par\smallskip
\centering
\includegraphics[width=.94\linewidth]{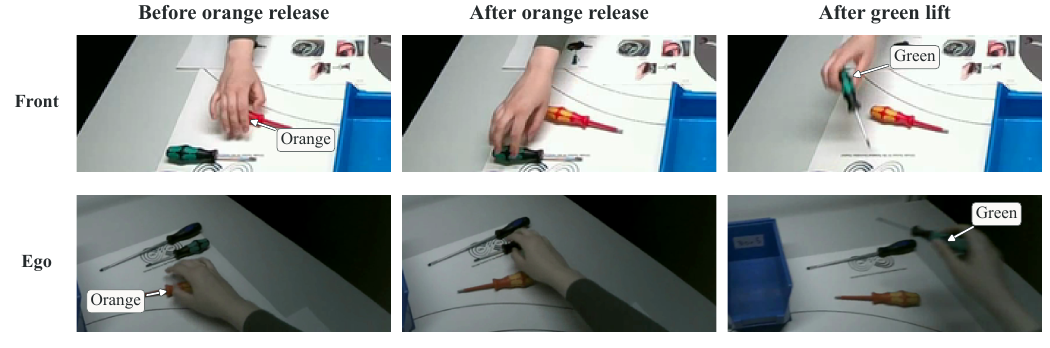}
\par\raggedright\small
\textbf{Figure F.1. Two views of a tool exchange.} The top row shows the front camera and the bottom row the first-person view. Columns align the visible stages while each view retains its own playback clock.
The start is the visible hand clearance from the orange tool, and the end is the visible lift of the green tool.
The ranges compare published playback coordinates for this external clip. This example illustrates agreement between two views of one event outside the eleven-source evaluation.
Cropped and marked IMPACT v1.1 TRAIN frames by Wen et al. (arXiv:2604.10409), CC BY-NC-SA 4.0.
\end{minipage}
\par\smallskip

\par\smallskip
\noindent\begin{minipage}{\linewidth}
\noindent\textbf{A window near the cutoff.}
In Fig.~F.2, use the tray rim to bracket the fingers' entry and withdrawal.

\par\smallskip
\centering
\hypertarget{osef-tray-example}{}\includegraphics[width=.94\linewidth]{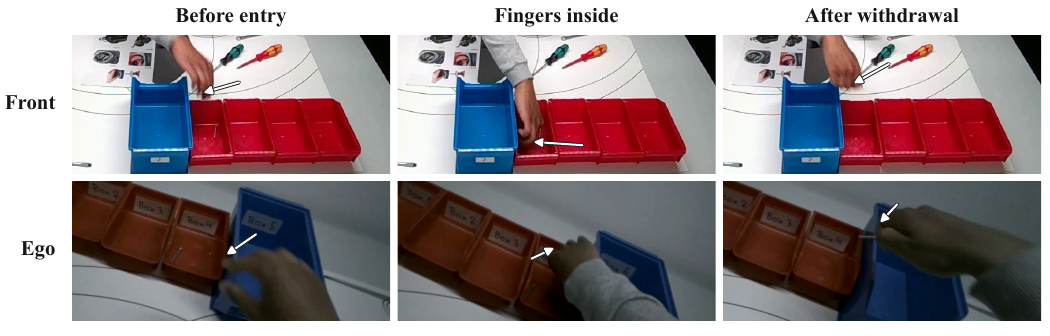}
\par\raggedright\small
\textbf{Figure F.2. Tray entry and withdrawal beside the blue bin.}
The rim and fingertips overlap in projection, so adjacent frames can leave the precise crossing unresolved. The columns match visible stages across views.
Cropped and marked external IMPACT v1.1 TRAIN frames by Wen et al. (arXiv:2604.10409), CC BY-NC-SA 4.0.
\end{minipage}
\par\smallskip

\par
\noindent\begin{minipage}{\linewidth}
\centering
\hypertarget{osef-ring-example}{}\includegraphics[width=.94\linewidth]{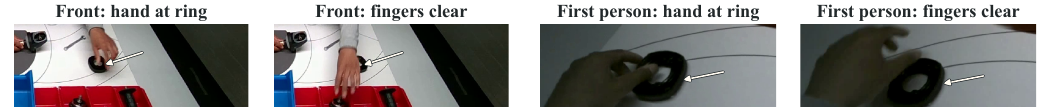}
\par\raggedright\small
\textbf{Figure F.3. A ring placed and released.}
The ring is flat before the fingers clear. Cropped and marked external IMPACT v1.1 TRAIN frames by Wen et al.\ (arXiv:2604.10409), CC BY-NC-SA 4.0.
\end{minipage}
\par\smallskip

\noindent\textbf{Three outcomes at the $0.5$ IoU cutoff.}
\par\smallskip
\begingroup\small
\begin{tabularx}{\linewidth}{@{}X l l l@{}}
\toprule
Action & Released-clock midpoints & Aligned midpoints & All aligned brackets \\
\midrule
Housing lift (Fig.~E.2) & 0.902 & 0.652--0.905 & 0.549--1.000 \\
Tool exchange (Fig.~F.1) & 0.000 & 0.708--0.833 & 0.320--1.000 \\
Tray reach (Fig.~F.2) & 0.337 & 0.622--0.947 & 0.49902--1.000 \\
\bottomrule
\end{tabularx}\par\endgroup
\smallskip\noindent
The lift passes across the audited range. The tray reach can miss by less than $0.001$. Clock alignment can rescue the tool exchange. Each fit uses four other visible events. See \path{data/VERIFY_IMPACT_CLOCK_EXAMPLE.py}.
\clearpage

\section*{Guide to the Supplement}
In Main Table~3(a), the six native settings use distinct-task pools. The eight converted diagnostics follow source-specific sampling rules (Sec.~A.2). The paired comparisons use development data that informed the study. Evidence paths are relative to the source root.

\begin{table}[ht]
\centering
\small
\begin{tabularx}{\linewidth}{@{}p{.25\linewidth}Xl@{}}
\toprule
Main paper & What to look up & Section \\
\midrule
Figs.~1--2, Tabs.~1--2 & \hyperref[app:compact-query-screen]{Fig.~2 audit and colors}, \hyperref[tab:compact-sources]{source inventory}, \hyperref[app:compact-linked-reference]{linked references} & A. Benchmark \\
Fig.~3, \hyperlink{osef-table3a-pools}{Tab.~3(a)}, Eqs.~(2)--(6) & \hyperref[app:compact-boundary]{Model}, \hyperref[tab:compact-donors]{Adapters}, \hyperref[app:compact-fig3-example]{\textbf{Fig.~3(b) invalid masses corrected}} & \hyperref[app:compact-method]{B. Model} \\
Tabs.~3--4, Fig.~4 & \hyperref[tab:compact-paired-decisions]{Tab.~3(b) paired counts}, \hyperref[app:compact-inputs]{Tab.~4 input controls}, \hyperref[app:compact-fig4-cases]{Fig.~4 saved cases} & \hyperref[app:compact-results]{C. Results}, \hyperref[app:compact-reference-sensitivity]{C.3}, \hyperlink{osef-timing-example}{F. Timing example} \\
Fig.~3, Secs.~5.2, 7 & Compute, evidence limits and \hyperlink{impact-visibility}{a frame example} & \hyperref[app:compact-scale]{D. Compute}, \hyperref[app:compact-limits]{E. Limits} \\
\bottomrule
\end{tabularx}
\caption{Where to find the details behind the main paper.}
\label{tab:compact-guide}
\end{table}

\section{Benchmark and Evaluation}\label{app:compact-benchmark}
\label{app:main-details-002}

\subsection{From source annotations to CVSPP examples}
\label{app:construction-notes-r23}\label{app:historical-coverage-086}
\label{app:task-display-notes-r57}\label{app:converted-construction-r057}

The model receives a start--goal query $q$, an ordered candidate list and the requested action count $T$. It returns a video index, a temporal interval and $T$ ordered action labels. The query omits fields for the construction video index, timestamps, target actions and intermediate states. The plan preserves annotation order. The hull in Main Sec.~3 applies to native sources. Converted intervals use the first start and last end (Sec.~A.3). See the \hyperref[app:compact-query-screen]{query audit} for Main Figure~2.

Main Table~2 counts eligible videos and constructed windows across TRAIN and VALID. Its eleven sources contribute 25,750 eligible videos and 192,493 windows at $T=3$. Main Table~3(a) evaluates both horizons on the three native sources and $T=3$ on the eight converted sources, for fourteen settings. The converted $T=4$ columns in Main Table~2 record construction counts. The taxonomy-based EgoExoLearn (EEL) conversion is a separate diagnostic.

\begin{table}[!htbp]
\centering\small
\setlength{\tabcolsep}{4pt}
\renewcommand{\arraystretch}{1.1}
\caption{\textbf{Source-specific construction and evaluation populations.}
Label codes indicate native (N), weak (W), released (R) and taxonomy-based (T)
targets. Native time bases and splits describe Main Table~3(b).
Converted rows describe the historical construction.}
\label{tab:compact-sources}\label{tab:converted-construction-r057}
\label{tab:historical-reference-coordinates-r172}\label{tab:candidate-construction-r191}
\begin{tabularx}{\linewidth}{@{}>{\raggedright\arraybackslash}p{.174\linewidth} >{\raggedright\arraybackslash}X >{\raggedright\arraybackslash}X >{\raggedright\arraybackslash}p{.185\linewidth}@{}}
\toprule
Source & Window and label & Time base for reference & Split and group key \\
\midrule
\multicolumn{4}{@{}l}{\textit{Native sources, $T=3,4$ in the adaptation inventory}}\\[1pt]
NIV~\cite{niv2018} &
Consecutive records in original list order (N). &
Hull over legacy feature-row count. &
Fixed split, task name. \\
COIN~\cite{coin2019} &
Consecutive annotated actions (N). &
Hull over stored annotation duration. &
Fixed mixed cohort, task name. \\
CrossTask (CT)~\cite{crosstask2019} &
Consecutive rows ordered by start time and source row index (N). &
Hull over maximum retained end. &
Video-ID hash split, task name. \\
\addlinespace
\multicolumn{4}{@{}l}{\textit{Converted sources, $T=3$ in the adaptation inventory}}\\[1pt]
YouCook2 (YC2)~\cite{youcook2_2018} &
Recipe step sentences (W). &
Step times over annotation duration. &
Project split, recipe type. \\
TACoS~\cite{tacos2013} &
Cooking sentences from OSGNet~\cite{osgnet2025} (W). &
Sentence times over annotation duration. &
Merged partitions, project split, dish code when eligible. \\
HiREST (HiR)~\cite{hirest2023} &
Step headings with a following step (W). &
Absolute steps over annotated video duration (zero if missing). &
Source train/val, instructional goal. \\
Ego4D GoalStep (GS)~\cite{ego4dgoalstep2023} &
Top-level descriptions, with at least $T+1$ procedural steps per video (W). &
Times relative to annotated start over positive duration or latest top-level end. &
Source train/valid, goal category. \\
ActivityNet Captions (ANet)~\cite{anetcaptions2017} &
VSLNet~\cite{vslnet2020} event sentences, tokens with at least three letters (W). &
Event times over annotation duration. &
Project split, no task key. \\
Charades-Ego (CEgo)~\cite{charadesego2018} &
Class descriptions for boundaries and codes for targets (R). &
Class times over CSV length (zero if missing or invalid). &
Merged source train/test, project split, scene. \\
GUIDE~\cite{guide2024} &
Guideline steps assigned to substeps (R). &
Substep times over latest end in last eligible video record. &
Project split, stripped guideline query. \\
EgoLearn (ELearn, project conversion)~\cite{egoexolearn2024} &
English egocentric narrations with task metadata (W). &
Narration times over frames/fps (1 fps if missing or zero). Use latest end for nonpositive duration. &
Project split, task metadata. \\
\addlinespace
\multicolumn{4}{@{}l}{\textit{Separate diagnostic outside the eleven-source totals}}\\[1pt]
EgoExoLearn (EEL)~\cite{egoexolearn2024} &
Positive-duration egocentric segments. The first matched word maps to a released verb (T). &
Greater of latest end and cached S3D length at a nominal second per row. &
Grouped split, scene and task. \\
\bottomrule
\end{tabularx}
\end{table}

Converted windows take $T$ consecutive steps in start order from
feature-matched videos with at least $T+1$ steps. Neighbors give start and
goal text or fixed defaults. HiREST needs a next step. Builders lowercase,
remove digits and normalize spaces. A weak label is the first alphabetic
token outside a stop list, or \texttt{do}.

Project splits shuffle sorted video IDs in NumPy and take the first
$\max(1,\lfloor0.2n\rfloor)$ of $n$ for VALID. Seeds are 20260614 (YouCook2,
TACoS) and 20260615 (ActivityNet Captions, Charades-Ego, GUIDE, EgoLearn).
EgoLearn uses VideoMAE. Input pins are in \path{data/DATA_SOURCE_PROVENANCE.txt}. \phantomsection\label{app:converted-eel-r063}EEL
uses S3D, resolves inflections and defaults to \texttt{<other>}. It samples
about 20\% per scene and task (seed 20260701), leaving four TRAIN videos
where possible. Archived S3D counts reproduce 51,859 reference windows for 177 videos. Logs report pre-write checks of the construction slot and I16 entry.
Publisher-derived row counts shift one constructor I16 endpoint in
3/5,314 ($T{=}3$) and 3/5,277 ($T{=}4$) fixed exposed VALID windows (\mbox{\path{data/eel_reference_projection.json}}).
The EEL candidate roster, alternate supports and original cache reads remain unverified.

\subsection{Candidate sampling and leakage checks}

\paragraph{Candidate construction.}
\hypertarget{osef-table3a-pools}{}
The main joint comparisons use $K=4$, with the construction video, two others from its task and one from another task. Candidate identity and order stay fixed across paired decisions. WV denotes another same-task video, WO a wrong window in the construction video, and WA a video from another group. A sampled alternative can have valid support. The six native Table~3(a) settings use \emph{distinct-task} candidates. These lists and the historical YouCook2, TACoS, HiREST and GoalStep lists ordinarily fill three WA slots. ActivityNet fills its WA slots without a task key. Charades-Ego and EgoLearn fill three WV slots, and GUIDE mixes them. Some historical builders repeat a candidate when the pool runs out.

\paragraph{Evaluation populations.}
The main NIV cohort uses 103 TRAIN and 45 exposed VALID videos, with 691/333 windows at $T=3$ and 589/288 at $T=4$. The historical 147 eligible NIV $T=3$ videos in Main Table~2 and the current 148 video identities describe separate constructions. CrossTask's current video-ID split takes hash buckets 0--14 for VALID and 15--99 for TRAIN and supplies 2,028 exposed VALID cases at $T=3$. COIN's mixed cohort has 14,558 TRAIN and 4,685 exposed VALID cases at $T=3$. The historical COIN $T=3$ inventory has 14,569 TRAIN and 4,700 VALID windows. Reported evaluations use exposed validation or training records. The historical adaptations can select checkpoints on VALID. The principal NIV fits fix epoch 100 in advance.

\paragraph{Reading the audit gate in Main Figure~2.}\label{app:compact-query-screen}
Fig.~2(a)'s redaction gate screens native queries. Converted queries may keep target words.
In mixed $T{=}3$ VALID, exact catalogue phrases remain in 0/2,028 CrossTask and 0/4,685 COIN queries.
Panel (b) shows TRAIN and VALID shares of 25,750 eligible videos. Historical vocabularies and COIN's lexicon span splits.
Candidate IDs are disjoint across splits, as are COIN file and selected-feature hashes. Respectively, 2,014 and 4,312 VALID queries
recur in TRAIN, whose modes match 1,199 and 3,193 canonical plans. For new queries with seen task-matched endpoints,
single-endpoint modes score 0/13 and 0/138. Joining endpoint pairs at a common middle action and ranking by count
products yields 3/13 and 49/138 (\mbox{\path{data/query_composition.json}}).
With identical query and task but different plans across videos, one deterministic plan per key matches at most
859/1,463 and 1,309/1,966 canonical plans, respectively. No pair in these audited cohorts differs solely in order (\mbox{\path{data/text_plan_ceiling.json}}).
Charades-Ego has cross-split pairs. Paraphrases, alternate support and cross-source overlap remain unverified.

\paragraph{What the colors in Main Figure~2 mean.}
\emph{Usable converted} (orange) marks HiREST and GoalStep plan diagnostics. Converted rows are excluded from method ranking.
The best donor on each brown-cross source gains 0--6 $VP$ hits over the fixed-retrieval TRAIN majority.
HiREST reaches 27 versus 25 of 377 (\mbox{\path{data/figure2_legend_scope.json}}).
The hit gains overlap across these labels.
GUIDE has 933/1,291 exact VALID query--plan pairs in TRAIN (72.3\%). Duplicate footage remains unverified.
EEL target-label phrases occur verbatim in 2,079/26,018 ($T=3$) and 2,640/25,841 ($T=4$) TRAIN and VALID queries
(\mbox{\path{data/eel_build_query_audit.json}}).

\subsection{One reference per success and normalized time}\label{app:compact-linked-reference}

\paragraph{Complete witnesses.}
A reference $(k,I,A)$ gives the candidate slot, interval and full ordered plan. Canonical scoring uses the construction triple alone. Main Table~3(b)'s finite policy adds retained triples from the fixed list with the same query and horizon. Additional valid answers may be unannotated.

For a prediction $(\hat k,\hat I,\hat A)$, define $v=[\hat k=k]$, $p=[\hat A=A]$ and $i_\tau=[\operatorname{IoU}(\hat I,I)\ge\tau]$ for one reference. The scores are
\[
\begin{aligned}
V &= [\exists\, (k,I,A)\in\mathcal R:\ v], &
VI_\tau &= [\exists\, (k,I,A)\in\mathcal R:\ v\land i_\tau],\\
P &= [\exists\, (k,I,A)\in\mathcal R:\ p], &
VP &= [\exists\, (k,I,A)\in\mathcal R:\ v\land p],\\
J_\tau &= [\exists\, (k,I,A)\in\mathcal R:\ v\land i_\tau\land p].
\end{aligned}
\]
Square brackets are indicators. See the \hyperlink{osef-order-example}{opening visual example} for the two valid orders and the worked failure cases. Each conjunction requires one witness. Historical FULL-SR is $VP$ and has no interval test. Rates average the binary flags over all declared cases. Invalid predictions receive zeros and malformed references invalidate evaluation. The default evaluation cutoff is $\tau=0.5$.

\paragraph{I16 and the underlying time bases.}
With $M=16$ cells, stored inclusive indices $(i,j)$ represent
$[i/16,(j+1)/16)$ on the normalized annotation axis. IoU divides intersection length by union length. For native reference hulls $(u,v)$ (main Sec.~3), use $i=\lfloor15u\rfloor$ and $j=\lceil15v\rceil$, clipped to $[0,15]$ and $[i,15]$. COIN uses binary64 and CrossTask exact rational arithmetic. Native finite search retains full triples for the exact query. See \mbox{\path{data/native_i16_reference.json}}.
With first start $s_1$ and last end $e_T$, converted windows use $i=Q(s_1)$ and $j=\max\{i,Q(e_T)\}$. Here $Q(t)=\operatorname{round}(15\,\operatorname{clip}(t,0,D)/D)$ for duration $D>0$ and zero otherwise.
Main Sec.~3 describes enclosure for native hulls. For converted windows, an earlier overlapping step can extend past $e_T$. Sec.~6.5 tests unsnapped annotations. Physical timing remains unverified.

\textbf{Distinct plans can share a grid interval.}
On exposed HiREST~\cite{hirest2023} validation reconstructed from public annotations, 113/377 three-step and 122/316 four-step windows share a video and I16 interval with another window that has a different full plan.
Within each video, unsnapped endpoints or adding the exact query to I16 distinguishes every window. See \mbox{\path{data/hirest_interval_aliases.json}}.

\section{Model and Decoding}\label{app:compact-method}

The assembly goal in \hyperlink{osef-order-example}{Fig.~A.1} fits both recordings, but the order depends on the evidence. In Main Figure~3(a), OSEF scores candidate intervals and their action plans before choosing an output. The requested length $T$ sets the
number of action steps and the $2T$ Before/After queries.

\subsection{From the query to interval probabilities}\label{app:compact-boundary}
Let $x=(q,X)$ contain the start--goal query $q$ and clip features $X$. We fix the source
action vocabulary $\mathcal V$ and let $\theta$ denote all learned
parameters in the boundary and planning branches. The main model uses
$K=4$ candidate videos with $M=16$ ordered clips each. The validity mask
$C_{km}$ marks observed clips, and this configuration requires all 64 clips
to be valid. For video $k$ and clip indices $0\leq i\leq j<M$, hypothesis
$h=(k,i,j)$ covers the half-open interval $I_h=[i/M,(j+1)/M)$. The legal
set $\mathcal H$ has $H=KM(M+1)/2=544$ hypotheses.

Frozen BERT-base-uncased~\cite{bert2019} encodes the entire query. The
attention-mask mean of the final-layer token states, including special
tokens and excluding padding, gives $u(q)\in\mathbb R^{768}$. The 768
coordinates are text features. Separate learned maps $f_s$ and $f_g$ turn
this same vector into start and goal features. The shared visual map $f_v$
places each clip feature into the same 64-dimensional space. Each map uses
a linear layer followed by LayerNorm. Write $\bar s_q$, $\bar g_q$, and
$\bar v_{km}$ for their outputs after normalization by
$\max(\lVert\cdot\rVert_2,10^{-12})$. The boundary scores and interval
probabilities are
\begin{align}
 s_{km}&=10\bar v_{km}^{\top}\bar s_q, &
 g_{km}&=10\bar v_{km}^{\top}\bar g_q,\\
 E_h&=s_{ki}+g_{kj}, &
 \pi_h&=\frac{\exp E_h}{\sum_{h'\in\mathcal H}\exp E_{h'}}.
 \label{eq:compact-boundary-law}
\end{align}
Separate start and goal scores allow endpoints in different clips to
define a high-scoring interval. The signed logit $E_h$ ranks hypotheses. The softmax for $\pi_h$ runs over all video--interval
hypotheses.
The scalar $c_{km}=E_{(k,m,m)}/2=(s_{km}+g_{km})/2$ carries clip-level
matching information to the planner. The scaled cosine bounds give
$|c_{km}|\leq10$.

\subsection{From clip tokens to action probabilities}
For each hypothesis $h=(k_h,i_h,j_h)$, the token-global evidence-to-plan
adapter retains all 64 clips. For each $X_{km}$ it adds five terms in a common
width $D=256$:
\begin{align}
 r^h_{km}
 &= (W_xX_{km}+b_x) + (W_c\widetilde c_{km}+b_c)
    +\psi_q(u(q))+p_m+W_\gamma\gamma(h,k,m),\\
 \gamma(h,k,m)
 &=\big[\mathbf1\{k=k_h\},\ (m-i_h)/M,\ (m-j_h)/M,\ (j_h-i_h+1)/M\big]^\top .
 \label{eq:compact-geometry}
\end{align}
Here $\widetilde c_{km}=\operatorname{clip}(c_{km},-20,20)$, which equals
$c_{km}$ for these boundary scores. The query map $\psi_q$ is a learned Linear--LayerNorm--GELU map, and $p_m$ is the fixed
sinusoidal position feature. The sine and cosine at index $r$ use argument
$(m+1)10000^{-2r/D}$ and amplitude $\sqrt{2/D}$ for $0\le r<D/2$, with
positions repeated across videos. The matrices $W$ and biases $b$ are
learned. The bias-free geometry map $W_\gamma$ starts at zero. Its input
gives video membership, signed offsets to the two included boundary clips,
and the interval length. Figure~3(a) writes this per-clip vector as $\gamma(h)$. The adapter
keeps all 64 tokens.

\begin{table}[!htbp]
\centering
\begin{tabularx}{\linewidth}{@{}>{\raggedright\arraybackslash}p{.22\linewidth}>{\raggedright\arraybackslash}p{.24\linewidth}>{\raggedright\arraybackslash}X@{}}
\toprule
Operation & Shape for one $h$ & Main NIV configuration \\
\midrule
Clip contextualization & $64\times256\to64\times256$ &
One Transformer encoder layer~\cite{vaswani2017}. \\
State extraction & $64\times256\to2T\times256$ &
Eight-head cross-attention. The $2T$ learned query tokens start with
SD $0.02$ normal initialization and each receives $\psi_q(u(q))$. \\
State-pair projection & $2T\times256\to2T\times512$\newline $\phantom{2T\times256}\to T\times256$ &
Project each latent state, concatenate the Before/After pair at each
step, and linearly map its 1,024 features to a 256-dimensional step. \\
Query modulation & $T\times256\to T\times256$ &
The separate planner query map $768\to256$ uses Linear--LayerNorm--GELU.
Its linear FiLM head gives 256 scales and 256 shifts, applying
$z_t\odot(1+\mathrm{scale})+\mathrm{shift}$ to step $z_t$~\cite{film2018}. \\
Planner and action head & $(T+1)\times256\to T\times48$ &
Prepend the 256-dimensional planner query token, apply two Transformer
encoder layers, and send the $T$ step outputs through one shared linear
action head. \\
\bottomrule
\end{tabularx}
\caption{Tensor path from the Figure~3(a) clip tokens to NIV action logits.
Encoder layers use eight heads, feed-forward width 512, GELU, and
post-normalization. Attention and Transformer dropout are zero, LayerNorm
uses $\epsilon=10^{-5}$, and linear maps have biases except the geometry map.}
\label{tab:compact-tensor-path}
\end{table}

The step pairs $R_h\in\mathbb R^{2T\times256}$ are learned Before/After
representations. The plan loss supplies their supervision. The planner
yields $Z_\theta(x,h)\in\mathbb R^{T\times C_a}$, where
$C_a=|\mathcal V|$. Each row normalizes over action classes at its own step:
\begin{equation}
 p_{h,t}(a)=\operatorname{softmax}(Z_\theta(x,h)_{t,:})_a,
 \qquad P_\theta(h,A\mid x)=\pi_h\prod_{t=1}^Tp_{h,t}(A_t),
 \quad A\in\mathcal V^T.
 \label{eq:compact-joint-model}
\end{equation}
Self-attention couples the step representations. Given $x,h$, the output
likelihood factors over positions. For each $h$, summing the product over all
complete plans gives $\prod_t\sum_a p_{h,t}(a)=1$. Summing over $h$ then
gives $\sum_h\pi_h=1$. These normalized probabilities express model
preferences. Calibration requires separate evidence.

\paragraph{Main NIV settings.}
Frozen InternVideo-Next-Large~\cite{wang2026internvideonext} gives 4,096
patch tokens of width 1,024 per clip. We take the FP32 mean, apply a fixed
bias-free signed Walsh projection to 512 dimensions, and select 16 specified rows of 64. The $T=3$ and $T=4$ models train 3,043,184 and 3,043,696
parameters. The latter adds two extraction queries. Original-order input controls use seeds 0--2, AdamW~\cite{loshchilov2019adamw} at $10^{-4}$,
microbatches of eight, 16-hypothesis chunks, FP32 MATH attention, and
gradient clipping at 1. We fix epoch 100 before validation.

\paragraph{Visual access and planner adaptation.}
\phantomsection\label{app:compact-adaptation}
The main NIV route (G-score) uses all clips with their scores. G-uniform
zeros $c_{km}$ before its affine projection, keeping $b_c$, the visual
tokens, geometry and interval probabilities. W-score zeros visual and score
values outside each proposed interval and masks those keys in encoder and
extraction attention, while evaluating all 544 hypotheses. An early
hard-selection planner commits to one interval before planning and reads
only its evidence. Query-only zeros visual values in both branches while
retaining the query, public geometry and learned biases. The main COIN
model averages latent pairs extracted from all clips and from clips inside
each hypothesis. The routes share parameters and mix before the state
projection. The main CrossTask comparison uses a CONQUER-derived
locator~\cite{conquer2021} and a window planner (Sec.~\ref{app:compact-feature-controls}).

Main Table~3(a) compares historical systems. The six native settings use distinct-task candidates.
The eight converted settings are diagnostic (Sec.~A.2). Nine video scene procedure planners share a frozen retrieved
video and window per case and train on CVSPP targets. The export rounds $T+1$ evenly spaced positions to clips and pairs adjacent features.
Shared-plan OSEF reads global candidate clips and reuses one plan across intervals within each case.
In five historical VALID settings, its $VP$ count exceeds the most frequent reference-plan count, an upper bound for any
dataset-wide fixed plan (\mbox{\path{data/table3a_fixed_plan.json}}).

PDPP and ActionDiffusion read dataset task labels from training caches and validation exports.
All 142 archived fits record the earliest evaluated epoch with maximal self-retrieved $VP$. See \mbox{\path{data/table3a_donor_fits.csv}}
for counts, selected epochs and runner coverage. On converted sources, ViterbiPlanNet, SCHEMA and PlanLLM use BERT
before/after prompts in local action-ID order, with three repeated slots per prompt. ActionDiffusion uses BERT action-name features.
Main Figure~3's separate P3IV~\cite{p3iv2022} adapter supplies the first Before and last After states (512 dimensions each).
Main Table~4(c) fixes the retrieved interval and varies the input: raw endpoint clips, endpoint tokens, or tokens across all clips.

\begin{table}[!htbp]
\centering\small
\begin{tabularx}{\linewidth}{@{}p{.25\linewidth}X@{}}
\toprule
Planner & What the CVSPP adapter supplies \\
\midrule
ViterbiPlanNet~\cite{viterbiplannet2026} & State sequence and indexed descriptions. The action path reads the two outer states and decodes with training transitions. \\
MTID~\cite{mtid2025} & Outer states for the diffusion planner to interpolate its intermediate features. \\
SCHEMA~\cite{schema2024} & State sequence and indexed descriptions. The action path uses the outer states and predicted task, then applies Viterbi decoding. \\
P3IV~\cite{p3iv2022} & Outer states for the memory-based planner. The adapter uses action cross-entropy and Viterbi decoding of a seeded pass. \\
KEPP~\cite{kepp2024} & Outer states to predict bounding actions and condition diffusion. A graph from training data connects the bounding actions. \\
PDPP~\cite{pdpp2023} & Outer states and task category to condition ten-step DDIM. \\
ActionDiffusion~\cite{actiondiffusion2025} & Outer states and task category for ten-step DDIM. Action-text features supply training noise. \\
PlanLLM~\cite{planllm2025} & State sequence for the Stage~1 Q-Former. Intermediate states can affect the encoded endpoints. The adapter uses Viterbi decoding. \\
Skip-Plan~\cite{skipplan2023} & Projected outer states and learned intermediate queries for a parallel action decoder. \\
\bottomrule
\end{tabularx}
\caption{Inputs to the nine historical adapters in main Table~3(a).
The donor rows share retrieved evidence within each setting.
Training and inference access in the separate P3IV control are described
with main Table~4(c) in Section~\ref{app:compact-results}.}
\label{tab:compact-donors}
\end{table}

\subsection{A shared probability model for training and decoding}
\label{app:compact-decoding}
For the OSEF joint fit, take the construction reference $(k^*,I^*,A^*)$ and let
$\mathcal S^*=\{h\in\mathcal H:k_h=k^*,\operatorname{IoU}(I_h,I^*)\geq.5\}$.
In terms of $\ell(h,A)=\log\pi_h+\sum_t\log p_{h,t}(A_t)$, the joint event
loss is
\begin{equation}
 \mathcal L=-\log\sum_{h\in\mathcal S^*}P_\theta(h,A^*\mid x)
            =-\operatorname{logsumexp}_{h\in\mathcal S^*}\ell(h,A^*).
 \label{eq:compact-training}
\end{equation}
The on-grid construction interval itself belongs to $\mathcal S^*$.
With finite logits and in-vocabulary labels, the event therefore has
positive probability. Let $\omega_h$ be the posterior probability of
$h$ given this training event. For $h\in\mathcal S^*$ it is
$\exp(\ell(h,A^*)+\mathcal L)$, and it is zero outside that set.
Writing the loss as a difference of the evidence log normalizer and
the log sum of accepted evidence--plan weights gives
\begin{equation}
 \left.\frac{\partial\mathcal L}{\partial E_h}\right|_{Z}
   =\pi_h-\omega_h,\qquad
 \left.\frac{\partial\mathcal L}{\partial Z_{h,t,a}}\right|_{E}
   =\omega_h\bigl(p_{h,t}(a)-\mathbf1\{a=A_t^*\}\bigr).
 \label{eq:compact-gradients}
\end{equation}
The first term sends gradients through intervals that start or end at a
clip. The second sends gradients through the action planner. In the
full model, both $s_{km}$ and $g_{km}$ also receive
$\tfrac12\,\partial\mathcal L/\partial c_{km}$ through clip-score tokens.
This couples training of the two branches. Inference uses the model
distribution without reference annotations.

\paragraph{Point decisions and compatible mass.}
For each $h$, its modal plan $A^{(h)}$ chooses the highest-probability class
at each position (smallest class index on ties). Joint maximum a posteriori
(MAP) decoding takes the $h$ that maximizes $\ell(h,A^{(h)})$, using the
first lexicographic $(k,i,j)$ at a tie. The evidence-first decision chooses
the maximum $E_h$ and uses its conditional mode. The matched evidence-first
readout uses the same bank of 544 conditionals.

Compatible mass (CM) applies the accepted-event structure of training to
a proposed output $(d,A)$, with $d=(k_d,i_d,j_d)$ in place of a training
reference:
\begin{equation}
 \mathcal N_d=\{h\in\mathcal H:k_h=k_d,\ \operatorname{IoU}(I_h,I_d)\geq\tau_{\rm dec}\},
 \qquad U(d,A)=\sum_{h\in\mathcal N_d}P_\theta(h,A\mid x).
 \label{eq:compact-cm}
\end{equation}
The main setting takes $\tau_{\rm dec}=0.5$. The evaluation cutoff
$\tau$ compares the chosen interval with an annotated witness.
Each summand uses one hypothesis for every action in the same complete
plan, even if that plan is nonmodal there. Compatibility neighborhoods can overlap. 

\phantomsection\label{app:compact-fig3-example}\textbf{Endpoint consistency.} Consider two overlapping $V_1$ intervals with
distinct start and end indices. The example masses $0.30$ and $0.29$ cannot
coexist with that geometry under Eq.~(3). Swapping endpoints gives two more valid intervals
on the same video, so the argument sums four distinct hypotheses. At fixed video $k$, Eq.~(3) gives
$\pi_{(k,i,j)}\propto \exp(s_{ki})\exp(g_{kj})$. Hence the cross-paired
marginals satisfy $\pi_{(k,i,j')}\pi_{(k,i',j)}=\pi_{(k,i,j)}\pi_{(k,i',j')}$.
Each shown joint mass lower-bounds its interval marginal. By the
arithmetic--geometric mean inequality, the four marginals sum to at least
\[0.30+0.29+2\sqrt{0.30\times0.29}\approx1.18>1.\]

\hypertarget{osef-cm-examples}{}\textbf{Two normalized examples.} These use shared-start $V_1$ intervals with
IoU $6/7$\ifdefined\PlatformMainPhase, as in Figure~3(b)\fi. In both synthetic banks of 544 hypotheses,
point MAP picks $V_2$ with joint mass $0.31$ (rounded). CM picks the same
$V_1$ plan in both. Changing the residual mass moves the selected inclusive
bins from $[2,7]$ to $[2,9]$. The upper three rows show the dominant
summands. The exact endpoints use the full bank. The last row shows
the selected interval and its compatible mass. The axis marks clip boundaries.
See \mbox{\path{data/figure3_math_receipt.json}} for the full-bank maxima.

\par\noindent\makebox[\linewidth][c]{\includegraphics[width=.98\linewidth]{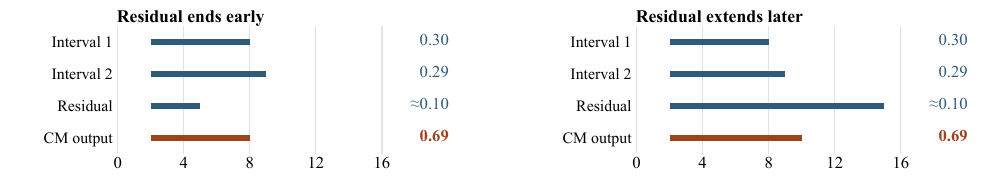}}\par

\paragraph{The plan pool and its guarantee.}
For each fixed input, deduplicate the complete modes $A^{(h)}$. At
$T=3$ the pool also includes the action tuple from the archived MAP decision, giving at most
545 plans. At $T=4$ the bank's own MAP returns a mode of one hypothesis,
so the native-mode pool contains it and has at most 544 plans.
Let $\mathcal P(x)$ denote this pool. Choose $(\hat d,\hat A)\in\arg\max_{d\in\mathcal H,\,A\in\mathcal P(x)}U(d,A)$.
Return its interval $I_{\hat d}$, video $k_{\hat d}$ and plan $\hat A$.
The readout uses binary64 log-softmax and log-sum-exp over $\mathcal N_d$. The tie rule applies among computed utility maximizers. Keep the point baseline if it is one of them. Otherwise use lexicographic interval and plan order. Matched decisions share their video list and conditional bank.

The decoder follows the minimum Bayes-risk principle used for
IoU-aware segmentation~\cite{nowozin2014optimal,ahmed2015optimizing}.
For one model-drawn triple, expected rejection is $1-U(d,A)$.
The MAP pair $(d_{\rm MAP},A_{\rm MAP})$ is feasible, so
$U(\hat d,\hat A)\ge U(d_{\rm MAP},A_{\rm MAP})$ for the same computed utility.

\textbf{Averaging the decoding cutoff.} Keep the proposed output fixed
and average its utility in Eq.~(\ref{eq:compact-cm}) over
$\tau_{\rm dec}\sim\operatorname{Unif}[a,b]$, with $0\le a<b\le1$.
A summand at IoU $r$ is accepted a fraction
$\min(1,\max(0,(r-a)/(b-a)))$ of the time. For example, IoU $0.5$ on
$[0.3,0.7]$ contributes half its mass. The benchmark uses its own
evaluation cutoff $\tau$.

Benchmark $J_{\tau}$ accepts an output whenever any complete reference
in $\mathcal R$ passes. It agrees in expectation with $U$ for one
model-drawn grid reference and equal cutoffs. For multiple references,
the union can favor a different output from a uniformly drawn member.
For example, fix the video and plan. With probability $0.6$ the set
consists of single-clip intervals at clips 0 and 7. Otherwise it contains
only clip 15. At cutoff $0.5$, no interval can match two of those clips.
Drawing one member favors clip 15 ($0.40$ versus $0.30$ per other clip).
Accepting any member favors clip 0 or 7 ($0.60$ versus $0.40$).

\begin{figure}[!htbp]
\centering
\includegraphics[width=.98\linewidth]{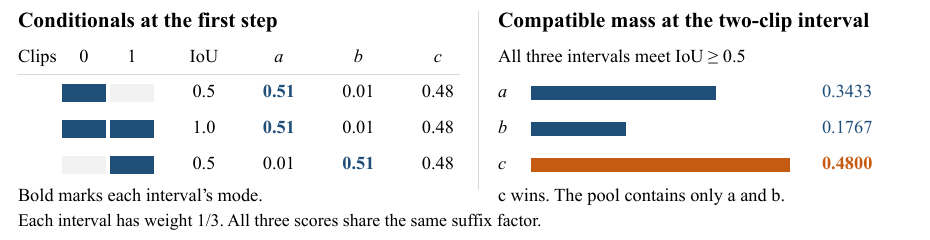}
\caption{An example where the modal plan pool misses the maximum.
At cutoff $0.5$, the two-clip output accepts all three intervals.
The bars show $U(d,A)$ divided by the common suffix probability.}
\label{fig:compact-pool-example}
\end{figure}

\paragraph{When can the pool miss a better plan?}
In Figure~\ref{fig:compact-pool-example}, three equal-weight intervals
share a modal suffix. The two-clip output receives every contribution.
It gives highest mass to the plan beginning with $c$, although every
individual mode begins with $a$ or $b$.

\FloatBarrier

\paragraph{Full-action checks.}
\phantomsection\label{app:compact-full-search}
For NIV and CrossTask~\cite{niv2018,crosstask2019}, split each plan at step 2 and sum compatible hypotheses by matrix product.
The pool matches exhaustive FP64 maxima and ties on 13,536 pairs (5,589 native NIV, 1,863 adapted NIV, 6,084 adapted CrossTask)
and 2,304 chronological-v2 NIV $T=4$ pairs. Point MAP with the same FP64 factors matches stored FP32 outputs on all 5,589 native
visual NIV pairs (\path{data/niv_map64_fixed_bank.json}), 6,084 CrossTask and 14,055 COIN pairs.

A prefix upper bound follows from Eq.~(\ref{eq:compact-cm}). For a plan prefix
$B=(B_1,\dots,B_r)$, each completion has $U(d,A)$ at most
\[
 \sum_{h\in\mathcal N_d}\pi_h\left(\prod_{t\le r}p_{h,t}(B_t)\right)
                                  \left(\prod_{t>r}\max_a p_{h,t}(a)\right).
\]
Prune below the incumbent. Retain equality when comparing tied outputs.

For COIN~\cite{coin2019}, main Table~3(b) uses the modal pool. Outward-bounded prefix search with exact dyadic factors
certifies 14,054 of 14,055 saved winners in the 778-label domain. One omitted plan wins under both scorers. Its
video and interval stay fixed, with no annotated witness on that video. The 16-bin grid and cutoff stay fixed.
These checks use reused development data. Sufficiency of the modal pool depends on the bank.

\section{Matched Experiments and Diagnostics}\label{app:compact-results}

\subsection{Which errors change under compatible-mass decoding?}
Main Table~3(a) scores $VP$ against the construction reference. Its six native settings use distinct-task negatives. See Section~\ref{app:compact-benchmark} for conversions.
Table~3(b) uses finite references and fixes models and banks (NIV scored global clips, COIN global/local states, and a CrossTask window planner with a CONQUER-derived locator~\cite{conquer2021}).

\begin{table}[!htbp]
\centering
\caption{Paired joint outcomes behind main Table~3(b), with the original
finite 16-bin references at IoU $\geq 0.5$. Gains and losses count transitions
on the same cases across three checkpoints. The final two columns partition
the gains whose MAP output already passed $VP$. The other gains came from
MAP outputs that failed $VP$.}
\label{tab:compact-paired-decisions}
\begin{tabular*}{\linewidth}{@{\extracolsep{\fill}}lrrrrrr@{}}
\toprule
 & & & \multicolumn{2}{c}{Joint transitions}
 & \multicolumn{2}{c}{MAP state before a gain} \\
\cmidrule(lr){4-5}\cmidrule(l){6-7}
Cohort & $N$/seed & Net gains by seed & Gains & Losses
 & $VP$, no $VI_{.5}$ & $VP$ and $VI_{.5}$ \\
\midrule
NIV $T=3$       & 333   & $+3,\ +8,\ +13$     & 66    & 42  & 53    & 0 \\
NIV $T=4$       & 288   & $+4,\ +18,\ +2$     & 45    & 21  & 36    & 0 \\
CrossTask $T=3$ & 2,028 & $+58,\ +61,\ +41$   & 289   & 129 & 205   & 20 \\
COIN $T=3$      & 4,685 & $+349,\ +318,\ +288$ & 1,543 & 588 & 1,091 & 42 \\
\bottomrule
\end{tabular*}
\end{table}

In 42 of COIN's CM-only gains, the MAP interval and plan matched different references. Both NIV outputs pass $P$
whenever joint status changes. Unsnapping CrossTask references changes the net contribution among positions where both
decoders pass $VP$ from $+147$ to $-45$, out of 6,084 positions in total (Sec.~C.3).

In Table~\ref{tab:compact-paired-decisions} order, SDs of $\Delta J_{.5}$ are 1.50/3.03/0.53/0.65 points.
By the paired margins, erasing the pooled gain needs at least 8/8/54/319 shared cases across all seeds excluded.\newline
See \mbox{\path{data/table3b_failure_path_r080.json}} and \mbox{\path{data/linked_witness_transitions_r081.json}}.

\textbf{Evidence-first versus joint MAP.} Both agree on finite NIV $J_{.5}$ flags (three seeds, 333/288 cases at $T=3/4$).
Evidence-first adds one canonical $T=4$ success with no loss (\mbox{\path{data/niv_evidence_first_pairs.json}}).

\textbf{Pool evidence, then plan.} Sum same-video evidence at IoU $\ge .5$ and read the chosen interval's modal plan
(288 NIV four-step cases in eight fits). This reproduces 2,080/2,097 Joint MAP-to-CM changes, with 19/2,304 differences from CM.
In the archive, 138 of 182 CM-only finite-reference $J_{.5}$ hits change only the interval.
If replayed MAP and CM match the scored tuples case by case, pooling retains at least 127 of these hits and nets 112--150 hits
over point evidence (\mbox{\path{data/niv_pool_gain_attribution.json}}, \mbox{\path{data/niv_pooling_replay.json}}).

\paragraph{A proposal-based locator.}
On exposed NIV with finite references, four QD-DETR~\cite{qddetr2023} fits per horizon give 1,332 and 1,152 fit--case pairs
at $T=3$ and $4$. Hard top-proposal selection gives 185 and 198 $VI_{.5}$ hits and 137 and 115 $J_{.5}$ hits.
On the same proposals, I16 projection and fitted planner, a mixture pools the softmax of all 40 log-foreground
scores by video and window and reads plan probabilities at each supported window. With this bank fixed,
MAP to CM changes joint hits from 230 to 267 and 190 to 225. One $T=3$ fit declines. Of the 130 and 113
CM-only joint hits, hard top proposal also misses 77--119 and 86--108, respectively (bounds from paired counts).
See \mbox{\path{data/qd_three_route_contrast.json}} for paired counts and fit-level variation.

\paragraph{Reading the saved cases in main Figure~4.}\label{app:compact-fig4-cases}
Both COIN cases output one video and one interval. They predict $[7,7]$
against $[7,8]$ in inclusive I16 cells (span IoU $0.50$). POS marks the
construction video, WV a same-task alternative, and blue boxes the reference
bins. Make Soap selects $V_1$ and matches the plan. Make Salmon selects $V_4$
over $V_2$ and differs at action 2, so $VI_{.5}$ and $J_{.5}$ fail. Its
$V_4$ clips 5 and 7 look alike at logits $1.32383$ and $1.33407$.
Orange labels give the softmax of the four video-max logits.
Soap $V_2$ ranks last at $0.179$. The orange bar in \hyperref[fig:case]{Figure~4} is scaled to that probability.
Hard selection fixes the evidence output. The historical planner reads all valid clips. Blue bars sum global endpoint attention
by video and average over six queries. Table~4(b)'s interval-only
control masks clips outside each interval in separate NIV fits.
See \path{data/FIGURE4_CASES_001.json} and \path{data/VERIFY_FIGURE4_001.py}.

\subsection{Which evidence and planner inputs matter?}\label{app:compact-inputs}
Main Table~4(a) compares separate NIV epoch-100 fits with shared initialization and training settings.
Query-only keeps query, interval geometry and biases, zeroing visual values. Video slots are symmetric in exact arithmetic
per span, so precision and ties can affect slot choice. Under MAP, query-and-video passes finite-reference $J_{.5}$
and query-only fails at 229/999 and 144/864 seed--case positions ($T=3,4$).
In at least 180 and 99, query-only passes $P$ but fails $VI_{.5}$. All three Table~4(b) visual inputs pass $J_{.5}$ in
at least 123 and 41 of these failures. In at least 51 and 20 gains, a single reference already supports the
query-only video and plan, but no reference on that video reaches IoU 0.5.
If query-only passed $J_{.5}$ on all queries absent verbatim from both TRAIN horizons, with all other outputs fixed,
query-and-video would still lead by 135 of 999 and 51 of 864 positions overall.
See \path{data/table4a_linked_bounds.json}, \path{data/niv_visual_support_overlap.json}, \path{data/niv_query_match.json}.

Separate fixed-epoch MAP fits show a similar distinction on CrossTask and COIN.
For CrossTask ($T=3$, $N=2{,}028$), adding video changes finite-reference $P$
by $+2.37\pm2.84$ percentage points and $J_{.5}$ by $+14.35\pm0.55$ points
across three matched seeds. For COIN ($T=3$, $N=4{,}685$), the three-seed
mean changes from $72.29\%$ to $70.00\%$ for $P$ and from $17.99\%$ to
$25.12\%$ for $J_{.5}$. These input controls use different fitted systems
from their respective Table~3(b) decoder comparisons.

\begin{table}[!htbp]
\centering
\caption{Paired evidence and clip-access diagnostics for NIV. Panel (a)
counts gains and losses in finite-reference $J_{.5}$ at the same
seed--case positions. Panel (b) changes planner access from the selected
interval to all 64 clips while keeping the indicated locator fixed.
Differences are in percentage points. The $\pm$ terms are sample SDs over
three paired seeds.}
\label{tab:compact-input-diagnostics}
\begin{tabular*}{\linewidth}{@{\extracolsep{\fill}}lcrrcr@{}}
\toprule
\multicolumn{6}{@{}l}{\textbf{(a) Privileged evidence with the same trained model}} \\
Change & $T$ & $N$/seed & New successes & Lost successes & $\Delta J_{.5}$ \\
\midrule
Native MAP to construction video  & 3 & 333 & 111 & 64 & $+4.70\pm2.72$ \\
Construction video to window  & 3 & 333 & 328 &  1 & $+32.73\pm0.79$ \\
Native MAP to construction video  & 4 & 288 &  99 & 23 & $+8.80\pm2.91$ \\
Construction video to window  & 4 & 288 & 190 &  1 & $+21.88\pm0.35$ \\
\addlinespace[3pt]
\multicolumn{6}{@{}l}{\textbf{(b) More planner clips with identical video and interval decisions}} \\
Locator query & $T$ & $N$/seed & $\Delta P$ & $\Delta VI_{.5}$ & $\Delta J_{.5}$ \\
\midrule
Full query tokens & 3 & 333 & $+3.30$ & $0$ & $-0.10\pm2.79$ \\
Pooled query token & 3 & 333 & $+1.90$ & $0$ & $+1.00\pm0.97$ \\
Full query tokens & 4 & 288 & $+3.82$ & $0$ & $+1.39\pm2.11$ \\
Pooled query token & 4 & 288 & $+2.55$ & $0$ & $+1.39\pm2.76$ \\
\bottomrule
\end{tabular*}
\end{table}

\paragraph{Supplying privileged evidence.}
Both Table~4(a) oracles reuse global scores and action logits, retain all clips and break ties in grid order.
The video oracle maximizes evidence in the construction video. The window oracle there maximizes IoU with the construction window.
Table~\ref{tab:compact-input-diagnostics}(a) shows losses when MAP succeeds on another witness.
Among the 18 three-action and 23 four-action cases without an exact query match in either TRAIN horizon,
window-oracle joint counts by seed are $(0,0,2)$ and $(0,1,1)$. Query-only $P$ is 0/54 and 3/69 without a match,
and 718/945 and 464/795 with one.

\paragraph{Clip access and fixed-locator controls.}
Main Table~4(b) compares separately trained input conditions. The
interval-only condition masks tokens outside each hypothesis. The no-score
condition zeros the planner's scalar clip input before projection. Both
score all 544 hypotheses. MAP and CM share the bank within each fit. The separate
CONQUER-derived study in Table~\ref{tab:compact-input-diagnostics}(b)
pairs selected-interval and 64-clip planner conditions with identical locator outputs.
Both conditions use the pooled planner query and epoch-100 checkpoints. With three actions and a full-token locator, broader planner
access improves mean plan accuracy while the mean joint rate falls
slightly. In seed 1, broader access adds two joint successes and loses
thirteen. This fixed-locator result distinguishes an improvement in the
standalone plan test from a plan that agrees with the selected evidence.

\paragraph{When input choices matter for decoding.}
With four actions, scored global and interval inputs both reach $21.06\%$ joint success
with CM (Table~4(b)). Each succeeds alone at 32 of 864 seed--case positions.
Let $D_a=J_{.5}(a,\mathrm{CM})-J_{.5}(a,\mathrm{MAP})$. Four paired tables per seed
give a sharp range for positions with different $D_a$: 25--129 of 864 for
these inputs, and 18--110 of 999 for scored and score-free global inputs
with three actions (\path{data/table4b_interaction_bounds.json}).
Integer outcome tables attain the summed seedwise bounds.
With three actions and CM, interval input leads by 9 of 999 with one
construction reference and global input leads by 5 with all recorded
references. Predictions and the grid are fixed
(\path{data/niv_input_reference_witnesses.json}).

\begin{figure}[!htbp]
\centering
\includegraphics[width=\linewidth]{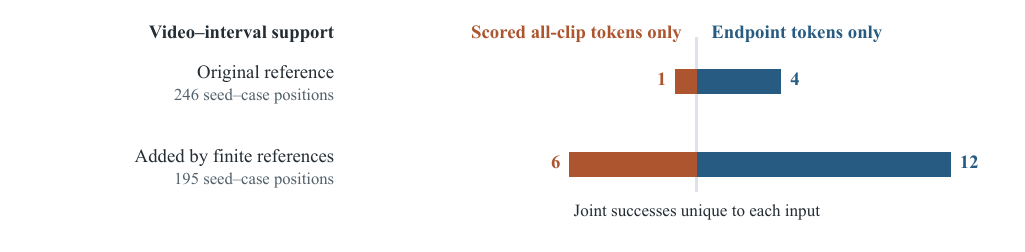}
\caption{Which P3IV input gets the whole joint prediction right?
Of 999 shared seed--case positions, the fixed locator passes $VI_{.5}$ on
246 with the construction reference and 195 more with the finite reference set.
Each bar counts finite-reference $J_{.5}$ successes unique to its planner input.}
\label{fig:compact-p3iv-pairs}
\end{figure}

\paragraph{Transferring to P3IV with fixed retrieval.}
Main Table~4(c) compares five input conditions, each trained at three
seeds on the separate chronological-v2 NIV $T=3$ split. Evaluation uses
the same 333 previously accessed validation cases per seed and fixed retrieval across conditions.
In \mbox{\emph{Predicted endpoints}}, P3IV~\cite{p3iv2022} reads the retrieved
window's raw boundary features in training and inference. Main Figure~3's separate
adapter supplies states in the four token arms. The \mbox{\emph{Endpoint-only tokens}}
arm keeps the two boundary tokens and their scores, masking other clips.
The other arms use all 64 clips with scores intact, zeroed, or shuffled.
Shuffling permutes valid-token mean scores within each example deterministically
by case and seed. All arms use probability-emission Viterbi with TRAIN-label transitions.
With finite references, all-clip scores yield 74 newly correct plans over zero scores.
Of these, 29 also pass the joint test, with no joint losses.
At least one other input also passes at 25 of the 29 positions. Against shuffled scores,
the joint-success differences are $-4,+2,+5$ by seed.
Compared with scored all-clip tokens, endpoint tokens gain 16 joint
successes and lose 7 (Figure~\ref{fig:compact-p3iv-pairs}). In these 23
one-arm joint successes, both arms pass the video--interval test. Only the winner's plan shares a reference with the fixed video and interval. Of the net gain of nine,
six come from positions whose fixed retrieved interval passes only when
scored against the finite reference set. Separate fits limit causal attribution.
Training-time retriever digests remain unverified for raw, score, and zero. See \mbox{\path{data/p3iv_table4c_archived_counts.json}},
\path{data/frozen_table4_diagnostics.json}, and \mbox{\path{data/p3iv_score_zero_controls.json}}.

\subsection{How sensitive are results to reference boundaries?}\label{app:compact-reference-sensitivity}
In NIV we fix candidates, predictions and each reference's video and plan.
Normalize the annotation hull $(s,e)$ by the legacy feature-row count $R$ and clamp $(u,v)=(s/R,e/R)$ to $[0,1]$.

\newpage\vspace*{-23pt}
\noindent\begin{minipage}{\linewidth}
The original grid (A) uses inclusive indices $\lfloor15u\rfloor$ and $\lceil15v\rceil$.
B uses the unsnapped $[u,v)$. The outward grid cover (D) uses $\lfloor16u\rfloor$ and $\lceil16v\rceil-1$.
Clamp grid indices to $0,\ldots,15$. Bin $m$ represents $[m/16,(m+1)/16)$.
\end{minipage}

\begin{figure}[!htbp]
\centering
\includegraphics[width=0.91\linewidth]{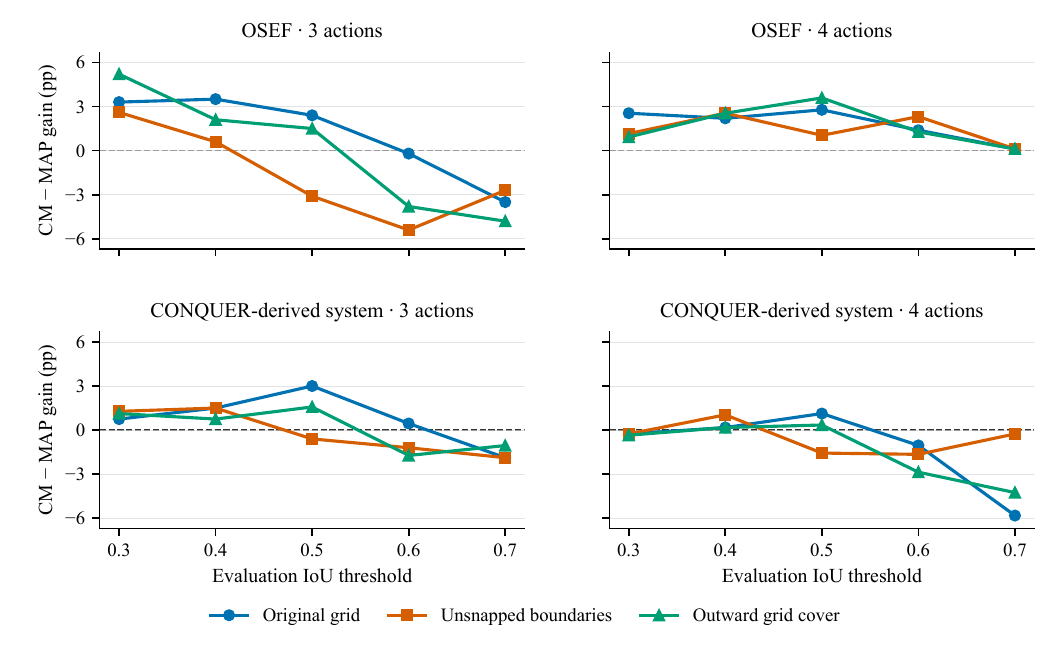}
\caption{Changing the evaluation reference can reverse the gain from
compatible-mass decoding. The curves show mean paired changes in joint
success on fixed NIV predictions. The evaluation IoU threshold varies along
the horizontal axis. All CM outputs were decoded at a compatibility
threshold of $0.5$. The plot uses three OSEF and four CONQUER-derived seeds
per horizon, with the same 333 or 288 development cases per seed.
The CONQUER-derived panels use a separate NIV replay with same-posterior
FP64 MAP. Curves show means across seeds.}
\label{fig:compact-reference-sensitivity}
\end{figure}

At $\tau=0.5$, B reverses three gains (Fig.~\ref{fig:compact-reference-sensitivity}).
For three-step OSEF, paired A outcomes and B totals bound drops in the per-case CM--MAP success gap
to 28--260 of 999 seed--case pairs. Equal A/B/D weights give $+0.27$ points ($0,0,+0.80$ by seed).
Nonnegative gaps across all seeds require B's weight $\le 1/3$.
Across five cutoffs $0.30$--$0.70$, equal weights over 15 reference--cutoff pairs give $-0.15$ points.
Keeping totals equal per reference and cutoff, means span $[-1.30,+1.13]$ points and the smallest seed gap peaks near $+0.06$.
% Keep the paths intact for PDF search and fit them in two short groups.
See \mbox{\path{data/niv_ref_bound.py}} and \mbox{\path{data/ref_cutoff_weights.json}}.\newline
\smash{Geometry counts are in }\mbox{\path{data/geometry_continuous_bounds.json}} and \mbox{\path{data/geometry_seed_counts.csv}}.

\noindent\begin{minipage}{.53\linewidth}
\textbf{Best possible grid overlap.}
Try both nearest grid edges per endpoint and keep the best positive-width interval.
Moving an overlapping output edge toward its reference edge without crossing
adds overlap or trims excess.
At $\tau=0.5$, reference length $1/2\le d<1$ cells requires a one-cell output
with overlap at least $a=(d+1)/3$. At a crossing the overlaps sum to $d$.
Both are below $a$ for $2a-d$ of each cell of starts, leaving fraction $(d+1)/3$.
\end{minipage}\hfill
\begin{minipage}{.45\linewidth}\centering
\includegraphics[width=\linewidth]{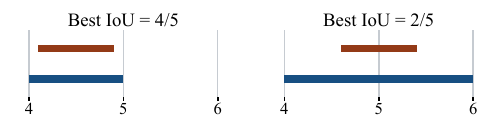}\par
{\small Two $0.8$-cell references (top) and their best grid outputs (bottom).
Across a full cell of starts, $60\%$ allow IoU $\ge0.5$.
Only the left example passes.}
\end{minipage}\par
\noindent At $\tau=0.5$, a reference at least one cell long always has a feasible grid interval.
For $1\le d<2$, use the two-cell cover or the full cell between two crossings.
For $d\ge2$, the cover adds at most two cells. \mbox{\path{data/grid_overlap_r396.json}}.

\textbf{Changing the decoding cutoff.} Raise $\tau_{\rm dec}$ from 0.5 to 0.9 with scores and finite I16 references fixed.
$J_{.5}$ drops (NIV 604 to 481, CrossTask 1,366 to 1,162) and $J_{.9}$ rises (NIV 8 to 46, CrossTask 90 to 188).
All 11 fits agree (2,304 NIV and 6,084 CrossTask validation fit--case pairs). \mbox{\path{data/paired_decoding_cutoffs.json}}.

For four-step NIV with finite references, MAP or CM (0.5) passes $J_{.5}$ in 656/2,304 fit--case pairs.
Adding CM at 0.8 and 0.9 gives 660--731 pairs with any pass in hindsight (\mbox{\path{data/niv_decoder_coverage.json}}).

\vspace{-.5\baselineskip}
\paragraph{More candidate intervals for the final decision.}
NIV replays 136 or 968 outputs per video with the same 544 latents, probabilities and plan pool.
At $\tau_{\rm dec}=0.5$, output $[s,e)$ of width $w$ supports latent $[a,b)$ of width $L$ if
$L/2\le w\le2L$, $s+2e\ge2a+b$ and $2s+e\le a+2b$.
Intersect a maximizer's closed compatibility constraints with the endpoint domain.
A feasible vertex retains all supported latents and possibly more, preserving maximum mass under nonnegative weights.
The expanded set adds valid intersections to the grid.
With both IoU cutoffs at 0.5, the smallest subsets of these 968 have 108 outputs to preserve which references can match
and 252 outputs to also preserve maximum compatible mass for any fixed nonnegative posterior.
The output selected in a tie can change (\mbox{\path{data/interval_subset.py}}).
Under B, grid and expanded CM each give 163/999 joint successes for three-action OSEF.
Per-seed totals imply at least seven gains and seven losses. From A to B, the pooled CM--MAP gap for the expanded outputs
moves from $+2.20$ to $-3.10$ points for OSEF and from $+2.70$ to $-0.83$ for the CONQUER-derived system.
Point MAP stays fixed. See \mbox{\path{data/niv_expanded_reference.json}}.

\vspace{-.5\baselineskip}
\paragraph{Other sources and candidate lists.}
Unsnapping Table~3(b)'s three-action references changes the CM gain from $+2.63$ to $-0.69$ points
on CrossTask and from $+6.79$ to $+3.63$ on COIN. CrossTask divides annotation endpoints by the retained
extent, and COIN by the stored duration. A separate CrossTask study uses three frozen checkpoints with
global scored clips on 2,028 development cases. Expanding each list from four to eight lowers $J_{.5}$ for both
decoders in all nine draw--checkpoint pairs, including references from added videos. CM drops by $0.40$ points less
on average (less in five pairs). Among positions where CM selected an added video, 29/565, 39/627, and 51/649
passed $J_{.5}$ across the three draws, respectively, pooling checkpoints per draw. See \path{data/crosstask_added_video.json}.

\FloatBarrier
\subsection{Joint success on converted-source controls}
Main Sec.~6.2 crosses eight converted sources, two heads (integer and public label embeddings), three
inputs (global clips, top-one evidence, and query plus structure with no visual values), and five seeds.
Each of the 240 fits runs for 120 epochs and selects the earliest maximum validation $J_{.5}$.
We score all five metrics on each full validation cohort using canonical 16-bin references.
The heads differ in capacity, which limits attribution to label representation.

The highest mean $J_{.5}$ across 48 settings is $2.467\%$ for YouCook2 with global clips and the integer
head (527 cases per seed). Converted labels can map distinct steps to the same plan, so matching a
plan gives limited evidence about the video or interval. The CSV \path{evidence/CONVERTED240_FIVE_SEED_FIVE_AXIS.csv}
gives source, head, input, seed, denominator and $V$, $P$, $VP$, $VI_{.5}$, $J_{.5}$ counts. Divide by the
denominator before averaging seeds. Main Table~3(a) uses video--plan success and its historical validation schedules.

\subsection{Feature and Oracle Controls}\label{app:compact-feature-controls}

\paragraph{Feature releases and model settings.}
NIV uses the encoders and Walsh projection in Sec.~\ref{app:compact-method}.
The CrossTask native input control and the documented COIN readout each sample
16 evenly spaced stored 512-D rows using rounded indices, without re-encoding
or averaging. CrossTask admits videos with at least 16 rows. Its native
control uses 133 classes and an equal global/local mixture. COIN input
controls use the frozen 768-D BERT-base query mean. The main COIN pair has
its own three epoch-100 fits, equal global/local mixture, and 778 classes.
The CrossTask and COIN feature producers remain unverified. A later COIN S3D
bank differs in row count and digest for all 63 overlapping training videos.
See \mbox{\path{data/coin_extractor_byte_check_r248.json}}.

\paragraph{CrossTask decoder bank.}
For Table~3(b), the CONQUER-derived locator~\cite{conquer2021} uses visual clips and one pooled BERT query token.
Start and end logits $S_{ki},G_{kj}$ give $E_h=S_{ki}+G_{kj}$ and
$\pi_h=\operatorname{softmax}_{h\in\mathcal H}(E_h)$ over 544 intervals.
The planner reads the query and window clips with zero scalar scores and uses Eq.~(\ref{eq:compact-joint-model}) for three $133$-class action distributions.
Training adds the two locator cross-entropies over 64 cells (weight $0.01$) to the construction-window plan's negative log probability.
Three AdamW fits use $10^{-4}$, batches of eight and epoch 100. MAP and CM share each bank.

\paragraph{Reversing visual feature inputs.}
On the same 333 exposed three-action NIV validation cases, reverse P3IV feature rows inside the fixed retrieved window, preserving video, window, query and validity mask.
With scalar scores fixed, probability-emission Viterbi changes plans on $4,0,0,1$ cases across four frozen aligned-query checkpoints and $46,0,56,23$ across four frozen query-permuted checkpoints.
Moving scores with rows changes no plan under probability-emission or raw-logit Viterbi.
These adapter tokens lack positional embeddings. See \mbox{\path{data/p3iv_feature_score_pairing_r075.json}}.

\paragraph{Complete oracle counts.}
NIV uses 333/288 VALID cases per seed at $T=3/4$ (Table~4(a--b)).
With finite references, positions passing only with the window oracle among MAP, CM and both oracles number
\mbox{207--274 of 999} and \mbox{128--174 of 864}. The video oracle loses 22--64 and 2--23 successes shared by both native decoders
(\mbox{\path{data/niv_oracle_bounds.py}}). Of 356/441 window-oracle failures, 299/411 match no reference plan,
56/30 match the plan only on another video, and 1/0 match video and plan but fail IoU at 0.5.
Changing only the interval cannot recover the first two groups.\newline
See \mbox{\path{evidence/oracle_paired_and_counts.tex}} and \mbox{\path{data/oracle_triplets_r214.json}}.\newline
CrossTask and COIN input-control tables are in\newline
\mbox{\path{evidence/crosstask_core_results_094.tex}} and \mbox{\path{evidence/coin12_full_results_047.tex}}.\newline
\textbf{Same construction plan.}
Among 795 exposed four-action NIV seed--case positions with a verbatim TRAIN query, both separately fitted arms
match the construction plan in 381. In that subset, G-score gains a net 29--80 construction $J_{.5}$ successes over query-only.
At least 34 G-score-only construction hits remain G-score-only with finite references. On the 31 positions where both also
choose the construction video, the net change in interval hits is $-10$ to $+27$.\newline
See \mbox{\path{data/niv_visual_attribution.json}}.

\section{Cost of Evaluating Candidate Intervals}
\label{app:compact-scale}
Global and interval-local extractors read $KM$ slots per interval. Shapes give 62.62 billion and 70.24 trillion
projection and attention multiply--accumulates per query at $K{=}4,100$ (544/13,600 intervals, \mbox{\path{data/niv_forward_macs.json}}).
On one generated $K{=}100$ input (random weights, one plan), 32 vs.~16 intervals per chunk cut median cache-to-CM time
by 15\%/17\% (3090/V100S) and keep decisions. PyTorch peak rises from 3.33 to 6.27 GiB (\mbox{\path{data/niv_chunk.json}}).
The exact CM log gap is $\ge 0.03369$ nats at width 16. Widths 8 or 32 can shrink it by at most $9.6{\times}10^{-6}$ nats
(\mbox{\path{data/cm_gap.json}}). CPU CM median on generated scores is 1.44/7.64 s (1/64 plans).\newline
See \mbox{\path{data/niv_cached_route_timing.json}} and \mbox{\path{data/niv_generated_forward_timing.json}}.

\begin{figure}[ht]
\centering
\includegraphics[width=\linewidth]{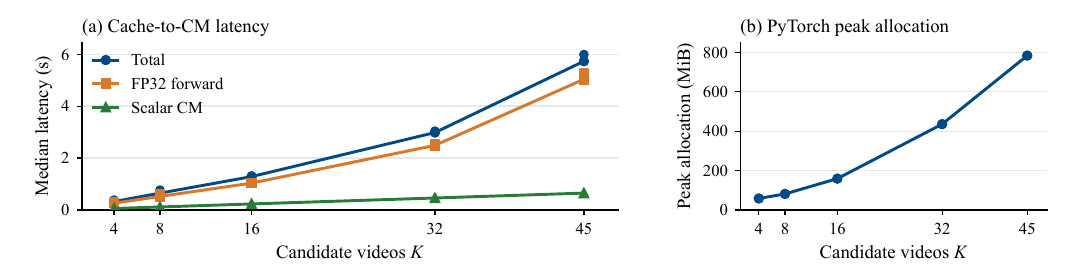}
\caption{At each $K$, eight frozen checkpoints reuse 288 exposed four-action NIV cases.
Marks span checkpoint medians or peaks on six RTX~3090 and two V100S GPUs. Lines connect minima.
From $K=4$ to $K=45$, median cache-to-CM time rises from 0.33--0.38 to 5.73--5.99\,s, and peak PyTorch allocation from 58.53 to 783.66\,MiB.
Timing excludes feature extraction, disk access, CM compatibility setup, warm-up and verification.
Joint accuracy below uses $K=4,6,8$.}
\label{fig:compact-scaling}
\end{figure}

Batch 1 uses FP32 chunks of up to 16 intervals. From $K=16$ to $45$, nominal forward MACs rise $9.80\times$.
Each checkpoint's median synchronized forward time rises $4.81$--$5.04\times$ (\mbox{\path{data/niv_dense_forward_scale.json}}).

In the first $K=8$ draw, peak same-video IoU-.5 evidence mass selects one pair (17/68 planner blocks).
Saved replay matches 2,298/2,304 full-bank CM tuples. With six changes unscored, archived paired counts bound
screened $J_{.5}$ to 571--581 vs.\ 576 full-bank hits on the same finite I16 references (\mbox{\path{data/niv_static_pair_screen.json}}).

\paragraph{NIV joint scoring.}
Using each pool's finite references, CM/point MAP $J_{.5}$ counts are 604/474 at $K=4$. The three draws give 597/468,
603/469, 582/465 at $K=6$ and 576/446, 590/471, 562/451 at $K=8$ (eight checkpoints; 2,304 fit--case positions per draw).
Fixing six-video references, expansion to $K=8$ costs CM 26--50 net hits. Adding references for the same outputs credits
13--30 further CM hits and flips point MAP's draw 1 change from $-18$ to $+2$. CM leads in all 48 state--draw comparisons
at $K=6,8$ on fixed four-video references. See \mbox{\path{data/niv_adjacent_pool_references.json}},
\mbox{\path{data/niv_nested_joint_counts.json}} and \mbox{\path{data/niv_same_pool_decoding.json}}.

\textbf{Candidate order.}
One prespecified reorder of each fixed $K=6,8$ set changes no video, interval or plan after aligning video IDs.
The 13,800 decoder comparisons cover evidence-first, point MAP and CM on 288 reused NIV four-action cases
at eight checkpoints, excluding identity orders. No reference labels are read. See \mbox{\path{data/niv_order.json}}.

\paragraph{NIV at 45 candidates.}
CM picks added videos in 1,633/2,304 pairs. With the original four videos' references, $V\le671/2304$ (1261/2304 at $K=4$).
With fixed reference choices, 137--234 of 288 cases gain a same-query witness on added videos at $K=45$.
In 17--213 cases, only $K=45$ adds a witness and none of the three $K=8$ draws does. See \mbox{\path{data/niv_reference_onset.json}}.\newline
See \mbox{\path{data/niv_k45_fixed_reference_bounds.json}} and \mbox{\path{data/niv_reference_availability.json}}.

\section{Limits of the Evidence}
\label{app:compact-limits}

\paragraph{Reference and data scope.}
Keep A primary and report B and D separately. Unsnapping can reverse gains. Physical time remains unverified.
On reused NIV T4 and CrossTask T3 with finite I16 references, averaging CM over cutoffs $0.5,0.8,0.9,1$
lowers $J_{.5}$ and raises $J_{.9}$ versus CM at $\tau_{\rm dec}=0.5$ (\mbox{\path{data/multicutoff_j_pairs.json}}).

\paragraph{Model comparisons.}
In Table~3(a), exact runner recovery covers 128/142 fits and 116/126 cells in full.
Process exits, actual inputs and provenance remain partly documented. See \mbox{\path{data/donor_runner_coverage_r322.json}}.

\paragraph{Sensitivity to evidence scale.}
\emph{Temperature replay.} On reused NIV T4 ($K=4$), vary $\eta$ in $\pi_h=\operatorname{softmax}_h(E_h/\eta)$ with fixed action probabilities.
Mean peak mass is $0.292/0.104/0.027$ at $\eta=0.5/1/2$. At $0.5$ and $2$ versus $1$, CM/MAP change 866/3 and 1,105/13 of 2,304 outputs.
At any one tested $\eta=0.5,1,2,4$, MAP changes at most 29 tuples. With fixed references, only those cases can gain hits.
If nominal tuples match the scored archives, subtracting 29 from the nominal CM leads of 57 and 130 gives
at least 28 more $J_{.5}$ successes on construction references and 101 more on finite references than each tested MAP arm
(\mbox{\path{data/niv_temperature_map_bound.json}}). Eq.~\eqref{eq:compact-training} can have negative curvature $\operatorname{Var}_{\pi}(E_h)-\operatorname{Var}_{\omega}(E_h)$ in $1/\eta$.
\emph{Gain control.} Recompute action probabilities with frozen model parameters. Seed~1 (cosine gain, $\eta$) pairs $(10,0.5)$ and
$(20,1)$ match $\pi_h$. Point MAP and evidence-first change plans in the same 51/288 G-score cases with video and interval fixed.
The zero-score G-uniform control is unchanged.
See \mbox{\path{data/niv_evidence_temperature.json}} and \mbox{\path{data/niv_frozen_gain.json}}.

\paragraph{Where performance remains open.}
Calibration, joint accuracy at $K=16,32,45,100$, pre-planner span filtering,
grounding, independent source tests and distractor validation remain open.

\par See the \hyperlink{impact-visibility}{opening occlusion example} for why visible seating cannot establish a physical completion time.

\FloatBarrier
\clearpage
\section{Fixed-Output Sensitivity}\label{app:compact-fixed-sensitivity}

\begin{figure}[!ht]
\centering
\includegraphics[width=0.95\linewidth]{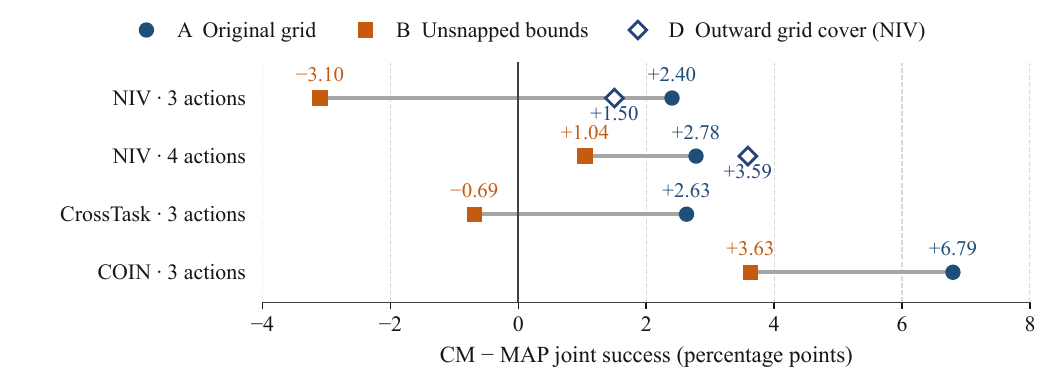}
\caption{Joint-success differences, CM minus MAP, with fixed outputs. A is the original I16 reference, B retains
the unsnapped endpoints and D is the outward I16 cover evaluated for the two NIV rows.}
\label{fig:compact-reference-all-four}
\end{figure}

\noindent\textbf{Reference boundaries in all four native comparisons.}
Figure~\ref{fig:compact-reference-all-four} pairs every Table~3(b) row at the same evaluation IoU of $0.5$.
The MAP and CM outputs, each reference's video and ordered plan, and CM's decoding cutoff $\tau_{\rm dec}=0.5$ stay fixed.
Removing the original I16 endpoint snapping reverses the gain for NIV three-step and CrossTask, and reduces
the gain for NIV four-step and COIN. The separate outward cover uses the NIV definitions in Sec.~C.3.
The three checkpoints in each row reuse their development cases.

\par See the \hyperlink{osef-timing-example}{opening two-view example} for the effect of playback coordinates and visible boundary uncertainty on IoU.

\begin{figure}[!ht]
\centering
\includegraphics[width=0.95\linewidth]{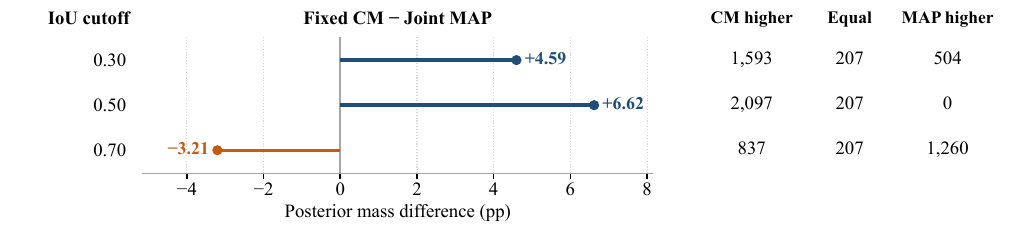}
\caption{The same CM output collects more mid-overlap mass and less high-overlap mass than MAP on average.
Over all 2,304 positions, the $[0.50,0.70)$ overlap band contributes $+9.825$ posterior-mass points and the
$[0.70,1]$ band contributes $-3.205$, for a total of $+6.620$ at $t=0.50$.}
\label{fig:compact-fixed-utility}
\end{figure}

\noindent\textbf{Overlap mass of the same two predictions.}
For a fixed prediction $y=(v,I,A)$, let $U_t(y)$ sum $P_\theta(h,A\mid x)$ over latent intervals $h$ in video $v$
with $\operatorname{IoU}(I,h)\ge t$. We compare the native MAP and $\tau_{\rm dec}=0.5$ CM outputs on the same
$K=4$ banks as the NIV four-action pooling diagnostic in Sec.~C.1. Eight frozen states reuse the same 288
development cases (2,304 paired positions). We check the original $U_{.5}$ exactly and evaluate the same two
outputs at $t=0.30,0.70$. Fig.~\ref{fig:compact-fixed-utility} measures model probability over latent triples.
Empirical $J$ uses the annotated reference set. Both outputs stay fixed at every mass cutoff.

\noindent Exact counts and plotting code ship with the supplement.\\
\mbox{\path{data/reference_points.json}}, \mbox{\path{data/utility_points.json}}.
\FloatBarrier

{\small
\bibliographystyle{ieeenat_fullname}
\bibliography{refs}
}

\end{document}